# The SEQBIN Constraint Revisited [*]

George Katsirelos[1], Nina Narodytska[2], and Toby Walsh[2]

[1] UBIA, INRA, Toulouse, France, email: george.katsirelos@toulouse.inra.fr
[2] NICTA and UNSW, Sydney, Australia, email: {nina.narodytska,toby.walsh}@nicta.com.au

**Abstract.** We revisit the SEQBIN constraint [1]. This meta-constraint subsumes a number of important global constraints like CHANGE [2], SMOOTH [3] and INCREASINGNVALUE [4]. We show that the previously proposed filtering algorithm for SEQBIN has two drawbacks even under strong restrictions: it does not detect bounds disentailment and it is not idempotent. We identify the cause for these problems, and propose a new propagator that overcomes both issues. Our algorithm is based on a connection to the problem of finding a path of a given cost in a restricted $n$-partite graph. Our propagator enforces domain consistency in $O(nd^2)$ and, for special cases of SEQBIN that include CHANGE, SMOOTH and INCREASINGNVALUE in $O(nd)$ time.

## 1 Introduction

Global constraints are some of the jewels in the crown of constraint programming. They identify common structures such as permutations, and exploit powerful mathematical concepts like matching theory, and computational techniques like flow algorithms to deliver strong pruning of the search space efficiently. Particularly eye-catching amongst these jewels are the meta-constraints: global constraints that combine together other constraints. For example, the CARDPATH meta-constraint [3] counts how many times a constraint holds down a sequence of variables. The SEQBIN meta-constraint was recently introduced in [1] to generalize several different global constraints used in time-tabling, scheduling, rostering and resource allocation. It also generalizes the CARDPATH constraint where the constraint being counted is binary. Our aim is to revisit the SEQBIN meta-constraint and give a new and efficient propagation algorithm.

## 2 Background

We write $D(X)$ for the domain of possible values for $X$, $lb(X)$ for the smallest value in $D(X)$, $ub(X)$ for the greatest. We will assume values range over 0 to $d$. A constraint is *domain consistent* (*DC*) if and only if when a variable is assigned any of the values in its domain, there exist compatible values in the domains of all the other variables of the constraint. Such an assignment is called a *support*. A constraint is *bound consistent* (*BC*) if and only if when a variable is assigned the lower or upper bound in its domain, there exist compatible values between the lower and upper bounds for all the other variables. Such an assignment is called a *bound support*. A constraint is *bounds disentailed* when there exists no solution such that each variable takes value between its lower and upper bounds. A constraint is *monotone* if and only if there exists a total ordering $\prec$ of the domain values such that for any two values $v, w$ if $v \prec w$ then $v$ can be replaced by $w$ in any support [5]. We define $\pi = (\pi^{bottom} := 0 \prec \ldots \prec d =: \pi^{top})$. A binary constraint is *row-convex* if, in each row of the matrix representation of the constraint, all supported values are consecutive (i.e., no two values with support are separated by a value in the same row without support) [6]. We use $x_{i,j}$ to represent the variable-value pair $X_i = j$. Let $C$ be a binary constraint. We write $(j, k) \in C$ if $C$ allows the tuple $(j, k)$. Consider a soft binary constraint $C$. We denote the cost of the tuple $c(j, k)$. If $(j, k) \in C$ then $c(j, k) = 0$ and $c(j, k) = 1$ otherwise. Given two sets of integers $S$ and $R$, we denote $S \uplus R = \{s + r \mid s \in S, r \in R\}$. Given a constant $c$, we write $S \uplus c = \{s + c \mid s \in S\}$. We denote $I[X]$ an instantiation of the variable sequence $X = [X_1, \ldots, X_n]$.

[*] NICTA is funded by the Australian Government's Department of Broadband, Communications, and the Digital Economy and the Australian Research Council. This work was partially funded by the "Agence nationale de la Recherche", reference ANR-10-BLA-0214.

## 3 The SEQBIN constraint

The SEQBIN meta-constraint ensures that a binary constraint $B$ holds down a sequence of variables, and counts how many times another binary constraint $C$ is violated.

**Definition 1.** *Given an instantiation $I[N, X_1, \ldots, X_n]$ and binary constraints $B$ and $C$, the meta-constraint SEQBIN$(N, X, C, B)$ is satisfied if and only if for any $i \in [1, n-1]$, $(I[X_i], I[X_{i+1}]) \in B$ holds, and $I[N]$ is equal to the number of violations of the constraint $C$, $(I[X_i], I[X_{i+1}]) \notin C$, in $I[X]$ plus 1.*

Note that we add 1 for consistency with the definition of SEQBIN in [1].

*Example 1.* Consider the SEQBIN$(N, [X_1, \ldots, X_7], C, B)$ constraint where $N = \{3\}$, $B$ is TRUE and $C(X_i, X_j)$ is a monotone constraint with one satisfying tuple $(1, 1) \in C$, $D(X_1) = D(X_3) = D(X_5) = D(X_7) = 1$ and $D(X_2) = D(X_4) = D(X_6) = \{0, 1\}$. Consider an instantiation $I[N = 3, X_1 = 1, X_2 = 0, X_3 = 1, \ldots, X_7 = 1]$. The constraint $C$ is violated twice: $(X_1 = 1, X_2 = 0)$ and $(X_2 = 0, X_3 = 1)$. Hence, the cost of the assignment is $N = 2 + 1 = 3$. □

A number of global constraints can be used to propagate SEQBIN including REGULAR [7,8], cost REGULAR [9], CARDPATH [3] and SLIDE [5]. However, all are more expensive than the propagator proposed here. A thorough analysis of related work is presented in [1]. We will assume that, as a preprocessing step, all binary constraints $B$ are made DC which takes just $O(nd)$ time for monotone $B$. We say that an instantiation $I[X]$ is $B$-coherent iff $(I[X_i], I[X_{i+1}]) \in B$, $i = 1, \ldots, n$. A value $v \in D(X_i)$ is $B$-coherent iff there exists a $B$-coherent instantiation $I[X]$ with $I[X_i] = v$.

### 3.1 A graph representation of SEQBIN

We present a connection between finding a solution of the SEQBIN constraint and the problem of finding a path of a given cost in a special $n$-partite graph where the cost of an edge is either 0 or 1. We start with a description of the graph $G(V, E)$. For each variable-value pair $x_{i,j}$ we introduce a vertex in the graph that we label $x_{i,j}$, $V = \{x_{i,j} | i = 1, \ldots, n, j \in D(X_i)\}$. For each pair $(x_{i,j}, x_{i+1,v})$ we introduce an edge iff the tuple $(j, v) \in B$, hence, $E = \{(x_{i,j}, x_{i+1,v}) | i = 1, \ldots, n-1, j \in D(X_i), v \in D(X_{i+1}) \wedge (j, v) \in B\}$. An edge $(j, v)$ is labeled with $c(j, v)$. Note that vertices $x_{i,j}, j \in D(X_i)$, form the $i$th partition as they do not have edges between them. Moreover, there are edges only between neighbor partitions $i$ and $i+1$, $i = 1, \ldots, n-1$. Hence, the resulting graph is a special type of $n$-partite graph that we call *layered*. To keep the presentation clear, we introduce dummy variables $X_0$ and $X_{n+1}$ with a single vertex $0^*$, and edges from $x_{0,0^*}$ to all vertices (values) of $X_1$ with cost 1, and from all vertices of $X_n$ to $x_{n+1,0^*}$ with cost 1. To simplify notation, we label a vertex $x_{i,j}$ that is at the $i$th layer simply as $j$ in all figures. We also use solid lines for edges of cost zero and dashed lines for edges of cost one. As variables correspond to layers in the graph we refer to layers and variables interchangeably. Similarly, as variable-value pairs correspond to vertices in the graph we refer to vertices at the $i$th layer and values in $D(X_i)$ interchangeably. Given two values $j$ at the $i$th layer and $v$ at the $(i+1)$th layer we say that $j/v$ is a support value for $v/j$ iff there exists an edge $(j, v)$ in the graph.

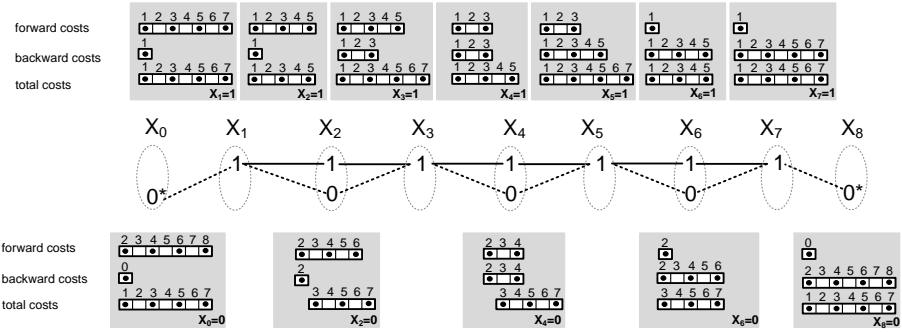

**Fig. 1.** A 9-partite graph that corresponds to the SEQBIN constraint from Example 1. Dashed edges have cost one and solid edges have zero cost.

*Example 2.* Consider the SEQBIN$(N, [X_1, \ldots, X_7], C, B)$ constraint from Example 1. Figure 1 shows the corresponding graph representation of the constraint. □

We now describe an algorithm, PATHDP to find a path of a given cost in a layered graph. PATHDP is a special case of the dynamic programming algorithm for the knapsack problem where all items have unit costs. Both the existing propagator for SEQBIN and our new one are specializations of PATHDP. Another specialization of PATHDP is the propagator for cost REGULAR [9]. We denote by $c(X)$ the set of all possible numbers of violations achieved by an assignment to $X$: $c(X) = \{k \mid I \text{ is } B\text{-coherent} \wedge c(I) = k\}$ and similarly $c(x_{i,j}) = \{k \mid I \text{ is } B\text{-coherent} \wedge I[X_i] = j \wedge c(I) = k\}$. We denote the *forward cost* from the variable $X_i$ to $X_n$ by

$c_f(x_{i,j}) = \{k \mid I[X_i, \ldots, X_n] \text{ is } B\text{-coherent} \wedge I[X_i] = j \wedge c(I) = k\}$. This set contains all the distinct costs that are achievable by paths from the vertex $x_{i,j}$ to the vertex $x_{n+1,0^*}$. We write $lb_f(x_{i,j}) = \min(c_f(x_{i,j}))$ and $ub_f(x_{i,j}) = \max(c_f(x_{i,j}))$. Similarly, we denote the *backward cost* from the variable $X_1$ to $X_i$ by $c_b(x_{i,j}) = \{k \mid I[X_1, \ldots, X_i] \text{ is } B\text{-coherent} \wedge I[X_i] = j \wedge c(I) = k\}$. It contains all the distinct costs that are achievable by paths from the vertex $x_{0,0^*}$ to the vertex $x_{i,j}$. We denote by $lb_b(x_{i,j}) = \min(c_b(x_{i,j}))$ and $ub_b(x_{i,j}) = \max(c_b(x_{i,j}))$.

---

**Algorithm 1** The pseudocode code for the PATHDP algorithm

---
1: **procedure** PATHDP( $G(V, E)$ )
2:    **for** $i = n \to 0; j \in D(X_i)$ **do**       ▷ Compute the forward cost
3:       $c_f(x_{i,j}) = \emptyset$
4:       **for** $k \in D(X_{i+1}), (j, k) \in B$ **do**
5:          $c_f(x_{i,j}) = c_f(x_{i,j}) \bigcup (c_f(x_{i+1,k}) \uplus c(j, k))$
6:    **for** $i = 1 \to n+1; j \in D(X_i)$ **do**       ▷ Compute the backward cost
7:       $c_b(x_{i,j}) = \emptyset$
8:       **for** $k \in D(X_{i-1}), (k, j) \in B$ **do**
9:          $c_b(x_{i,j}) = c_b(x_{i,j}) \bigcup (c_b(x_{i-1,k}) \uplus c(k, j))$
10:   **for** $i = 0 \to n+1; j \in D(X_i)$ **do**       ▷ Compute the total cost
11:      $c(x_{i,j}) = c_f(x_{i,j}) \uplus c_b(x_{i,j})$

---

PATHDP performs two scans of the layered graph, one from $X_n$ to $X_1$ to compute forward costs, and one from $X_1$ to $X_n$ to compute backward costs. The backward pass processes one layer at a time and computes the set $c_f(x_{i,j})$ for each variable $X_i$ and value $j \in D(X_i)$ (lines 2–5). Dually, the forward pass computes for each variable $X_i$ and value $j \in D(X_i)$, the backward cost $c_b(x_{i,j})$(lines 6–9). Finally, for each vertex the set of costs achievable on paths from $x_{0,0^*}$ to $x_{n+1,0^*}$ that pass through $x_{i,j}$ is $c_f(x_{i,j}) \uplus c_b(x_{i,j})$. To match the semantics of SEQBIN, we compute $c_f(x_{i,j}) \uplus c_b(x_{i,j}) \uplus (-1)$ for each vertex.

The time complexity for SEQBIN using PATHDP is $O(n^2 d^2)$ : the number of distinct costs is at most $n$, so getting the union of two cost sets takes $O(n)$ time. Each vertex has at most $d$ outgoing edges, so the set $c_f(x_{i,j})$ can be computed in $O(nd)$ time for each $x_{i,j}$. There are $O(nd)$ vertices in total, giving the stated complexity of $O(n^2 d^2)$.

*Example 3.* Consider the SEQBIN$(N, [X_1, \ldots, X_7], C, B)$ constraint from Example 1. Figure 1 shows the forward cost $c_f(x_{i,j})$, the backward cost $c_b(x_{i,j})$ and the total cost $c_f(x_{i,j}) \uplus c_b(x_{i,j}) \uplus (-1)$, $j \in D(X_i)$, $i = 0, \ldots, 8$ in gray rectangles. We have one rectangle for each variable-value pair $X_i = j$. Consider, for example, the vertex '1' at layer $X_5$. We compute the forward cost $c_f(x_{5,1}) = (c_f(x_{6,0}) \uplus c(1,0)) \cup (c_f(x_{6,1}) \uplus c(1,1)) = \{3\} \cup \{1\} = \{1,3\}$ and the backward cost $c_b(x_{5,1}) = (c_b(x_{4,0}) \uplus c(0,1)) \cup (c_b(x_{4,1}) \uplus c(1,1)) = \{3,5\} \cup \{1,3\} = \{1,3,5\}$. Then $c_f(x_{5,1}) \uplus c_b(x_{5,1} \uplus (-1) = \{1,3\} \uplus \{1,3,5\} \uplus (-1) = \{1,3,5,7\}$. □

**Lemma 1.** *Let $G(V, E)$ be a layered graph constructed from the* SEQBIN$(N, X, C, B)$ *constraint as described above. There exists a bijection between $B$-coherent assignments $I[X]$ of cost $s$ and paths in the graph $G(V, E)$ of cost $s + 1$.*

### 3.2 Revisiting SEQBIN

A domain consistency algorithm for the SEQBIN$(N, X, C, B)$ constraint, SEQBINALG was proposed in [1] under the restriction that $B$ is a monotone constraint. In this section we identify two drawbacks of this algorithm that make it incomplete. We show that SEQBINALG does not detect bounds disentailment and it is not idempotent even if $B$ is a monotone constraint. It was observed idependently in [10] that SEQBINALG does not enforce DC. However, the authors do not explicitly explain the source of the problems of SEQBINALG and only identify a very restricted class of SEQBIN instances where SEQBINALG does enforce DC.

We will identify the main reason that SEQBINALG fails to enforce DC. This is important to develop a new algorithm that does enforce DC in $O(nd^2)$ time when $B$ is monotone. SEQBINALG uses Algorithm 1 to compute only the lower and upper bounds of the forward and backward cost (Lemma 1 and 2 in [1]). Namely, using the notations in [1], we compute $\underline{s}(x_{i,j}) = lb(c_f(x_{i,j}))$, $\overline{s}(x_{i,j}) = ub(c_f(x_{i,j}))$, $\underline{p}(x_{i,j}) = lb(c_b(x_{i,j}))$ and $\overline{p}(x_{i,j}) = ub(c_b(x_{i,j}))$ in $O(nd^2)$. SEQBINALG is based on these values and runs in 4 steps [1]:

**Phase 1** Remove all non $B$-coherent values in $D(X)$.
**Phase 2** For all values in $D(X)$, compute $\underline{s}(x_{i,j}), \overline{s}(x_{i,j}), \underline{p}(x_{i,j})$ and $\overline{p}(x_{i,j})$.
**Phase 3** Adjust the min and max value of $N$ with respect to $\underline{s}(X)$ and $\overline{s}(X)$.
**Phase 4** Using the result of Phase 3 and Proposition 4 [1], prune the remaining B-coherent values.

The correctness of SEQBINALG relies on Proposition 3. Unfortunately, this proposition is not correct, and the algorithm is consequently incomplete.

**Proposition 3 (in [1]).** *Given an instance of* SEQBIN$(N, X, C, B)$ *with monotone $B$,* SEQBIN$(N, X, C, B)$ *has a solution iff* $[\underline{s}(X), \overline{s}(X)] \cap N \neq \emptyset$ *where* $\underline{s}(X) = \min_{j \in D(X_1)} \underline{s}(x_{1,j})$ *and* $\overline{s}(X) = \max_{j \in D(X_1)} \overline{s}(x_{1,j})$.

*Issue 1. Bounds disentailment.*

**Lemma 2.** *The algorithm* SEQBINALG *for* SEQBIN$(N, X, C, B)$ *with monotone $B$ does not detect bounds disentailment.*

*Proof.* Consider the SEQBIN$(N, [X_1, \ldots, X_7], C, \text{TRUE})$ constraint in Example 1. The constraint TRUE is monotone. Consider $N = 4$. Then $\underline{s}(X) = 1$ and $\overline{s}(X) = 7$. Hence, $[1, 7] \cap \{4\} \neq \emptyset$. However, there is no solution with cost 4. The problem with the proof of Proposition 3 in [1] is the last sentence which claims that there is a solution for each value $k \in [\underline{s}(x_{1,v}), \overline{s}(x_{1,v})]$ for some $v$. This is not true as Example 1 demonstrates. Note also that $[\underline{s}(x_{i,j}) + \underline{p}(x_{i,j}), \overline{s}(x_{i,j}) + \overline{p}(x_{i,j})] \cap \{4\} \neq \emptyset$. Hence, according to Proposition 4 [1] each variable-value pair is DC which is also incorrect. □

*Issue 2. Idempotency.* As a consequence of not detecting bounds disentailment, SEQBINALG is also not idempotent.

**Lemma 3.** *The filtering algorithm* SEQBINALG *for* SEQBIN$(N, X, C, B)$ *with monotone $B$ is not idempotent.*

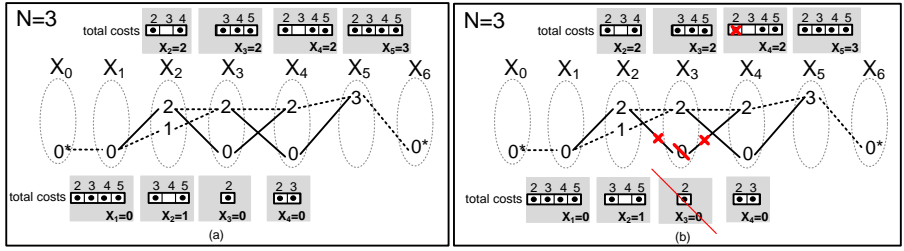

**Fig. 2.** A 7-partite graph that corresponds to the SEQBIN constraint from the proof of Lemma 3. Dashed edges have cost one and solid edges have cost zero. (a) shows initial costs; (b) shows costs after $X_3 = 0$ is pruned.

*Proof.* Consider the $\text{SEQBIN}(N, [X_1, \ldots, X_7], C, B)$ where $N = \{3\}$, $B = \{(j,k) | j, k \in [0,3], (j,k) \notin (0,0), (1,0)\}$. $C(X_i, X_j)$ is a monotone constraint with three satisfying tuples $(2,0), (0,2), (0,3) \in C$. Finally, $D(X_1) = \{0\}, D(X_2) = \{1, 2\}, D(X_3) = D(X_4) = \{0, 2\}$ and $D(X_5) = \{3\}$. Figure 2(a) shows the graph representation of the example. Note that $c(x_{3,0}) \cap N = \{2\} \cap \{3\} = \emptyset$. Hence, the value 0 is pruned from $D(X_3)$. Therefore, the value $X_4 = 2$ loses its support with cost 2 (Figure 2(b)). The new cost of $x_{4,2}$ is $\{4, 5\} \cap N = \emptyset$ and the value 2 is pruned from $D(X_4)$. Note that the removal of $X_4 = 2$ triggers further propagation as $X_2 = 2$ loses its support of cost 5, and 2 is removed from $D(X_2)$ at the next step. □

We note that if $B$ is not monotone, SEQBINALG may need $O(n)$ iterations to reach its fixpoint and Proposition 2 in [1] only works if $B$ is monotone.

*Remedy for* SEQBINALG. As seen in Lemmas 2–3, the main cause of incompleteness in SEQBINALG is that the set of costs for each vertex is a set rather than an interval even when $B$ is monotone. One way to overcome this problem is to restrict $\text{SEQBIN}(N, X, C, B)$ to those instances where it is an interval. This approach was taken in [10] where $\text{SEQBIN}(N, X, C, B)$ was restricted to *counting-continuous* constraints.

**Definition 2.** *The constraint* $\text{SEQBIN}(N, X, C, B)$ *is* counting-continuous *if and only if for any instantiation $I[X]$ with $k$ stretches in which $C$ holds, for any variable $X_i \in X$, changing the value of $X_i$ in $I[X]$ leads to $k$, $k + 1$, or $k - 1$ violations.*

This restriction ensures that the structure of the cost for each variable-value pair is an interval and, indeed, the filtering algorithm SEQBINALG enforces DC. However, this approach has a number of drawbacks. First, restricting $\text{SEQBIN}(N, X, C, B)$ to counting-continuous with monotone $B$ excludes useful combinations of $B$ and $C$. Example 1 shows that $\text{SEQBIN}(N, X, C \text{ is monotone}, B \text{ is TRUE})$ does not satisfy this property. Secondly, many practically interesting examples [1] that can be propagated in $O(nd)$ time do not satisfy these conditions. As was observed in [10], constraints $\text{CHANGE}^{\{=, \neq\}} = \text{SEQBIN}(N, X, C \in \{=, \neq\}, \text{TRUE})$ and SMOOTH are not counting-continuous. The INCREASINGNVALUE constraint which is $\text{SEQBIN}(N, X, =, \leq)$ violates the condition that $B$ is monotone. The only remaining constraint that satisfies these restrictions on $B$ and $C$ is $\text{CHANGE}^{\{<,\leq\}} = \text{SEQBIN}(N, X, C \in \{<, \leq\}, \text{TRUE})$. Unfortunately, the proof relies on the claim that $C$ is monotone, which is false for $C \in \{<, \leq\}$. Thirdly, we do not currently have a test to check if $\text{SEQBIN}(N, X, C, B)$ is counting-continuous. Despite the problems pointed out above, the filtering algorithm SEQBINALG enforces DC on INCREASINGNVALUE and $\text{CHANGE}(C \in \{<, \leq\})$ in $O(nd)$ as the counting-continuous property together with the row and column convexity of $C$ are sufficient to achieve this complexity.

In this work we take a different approach. We focus on an extension of the algorithm to handle non-interval cost sets. The challenge is to perform this extension in $O(nd^2)$ as the generic dynamic programming algorithm PATHDP that handles sets natively runs in $O(n^2d^2)$ time. Note that if the cost structure is an unrestricted set of values then the time complexity of PATHDP is going to be hard to improve as it is a specialization of a well-studied dynamic programming algorithm for the knapsack problem where all items have unit cost. Hence, we show that the structure of the costs for a variable-value pair is restricted if $B$ is monotone. This allows us to perform union operations on sets in $O(1)$ time rather than $O(n)$.

### 3.3 Cost structure

We show that the structure of the cost for each variable-value pair is restricted. First, we introduce definitions to formalize the structure of forward and backward costs.

**Definition 3.** *A set $S$ is a zipper set if it can be obtained from an interval $[a, b]$ by removing all odd or all even values. We denote a zipper set as $[a \sim b]$.*

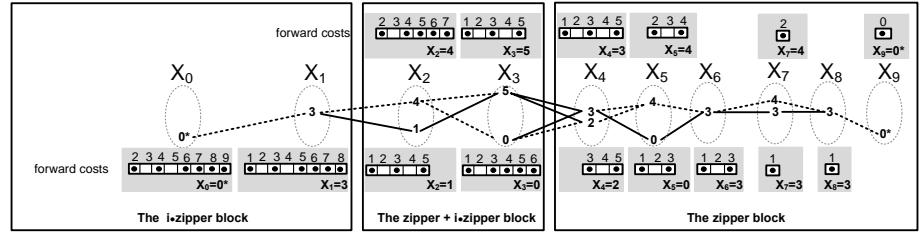

**Fig. 3.** A 10-partite graph that corresponds to the SEQBIN constraint from Examples 4–6. Dashed edges have cost one and solid edges have zero cost.

Note that in a zipper set $[a \sim b]$, $a$ and $b$ have the same parity. If both are odd, $[a \sim b]$ is an *odd zipper set*, while if both are even, $[a \sim b]$ is an *even zipper set*.

**Definition 4.** *A set $S$ is an i·zipper set if it can be written as $[a \sim b] \cup [b, c] \cup [c \sim d]$, $a \leq b < c \leq d$. We denote an i·zipper set as $[a \sim b - c \sim d]$. If $a = b$, we write the set as $[b - c \sim d]$ and if $c = d$ we write it as $[a \sim b - c]$.*

Given an i·zipper set $[a \sim b] \cup [b, c] \cup [c \sim d]$, we denote the left part $[a \sim b]$ as $l \cdot zip$, the middle part $[b - c]$ as $i \cdot val$ and the right part $[c \sim d]$ as $r \cdot zip$.

*Example 4.* Consider the SEQBIN$(N, [X_1, \ldots, X_8], C, B)$ constraint that Figure 3 presents. We only show the forward cost sets. For example, the forward cost set $c_f(x_{3,5})$ is a zipper $[1 \sim 5]$, $c_f(x_{5,4}) = [2 \sim 4]$ is an even zipper and $c_f(x_{3,5}) = [1 \sim 5]$ is an odd zipper. An example of an i·zipper set is $c_f(x_{1,3}) = [1 \sim 5 - 8]$. □

Our filtering algorithm is based on the following theorem.

**Theorem 1.** *Consider a SEQBIN$(N, X, C, B)$ constraint with monotone $B$ and arbitrary $C$. Let $[b, c]$, $c > b$ be the maximal interval such that $[b, c] \subseteq c_f(x_{i,v})$, $i = 1, \ldots, n, v \in D(X_i)$. If such an interval does not exist we define $[b, c] = \emptyset$. Then the following holds for any value $j, k, \{j, k\} \in D(X_i)$ and $i = 1, \ldots, n$:*

1. **Uniqueness.** *The set $c_f(x_{i,v})$ is either a zipper or i·zipper set.*
2. **Overlapping.** *If $c_f(x_{i,j})$ and $c_f(x_{i,k})$ are i·zipper sets, $c_f(x_{i,j}) = [a \sim b - c \sim d]$ and $c_f(x_{i,k}) = [s \sim r - q \sim t]$, then $[b, c] \cap [r, q] \neq \emptyset$.*
3. **Structure.**
   - Bounded holes. *If $c_f(x_{i,j})$ is an i·zipper set, $[a \sim b - c \sim d]$ then $b - a \leq 4$ and $d - c \leq 4$*
   - Closeness. $|lb_f(x_{i,j}) - lb_f(x_{i,k})| \leq 2$ *and* $|ub_f(x_{i,j}) - ub_f(x_{i,k})| \leq 2$.

Theorem 1 shows that the structure of $c_f(x_{i,j}), j \in D(X_i)$ is limited to few distinct structures of sets: a zipper and an i·zipper. This allows us to deal with such restricted sets efficiently. We give an overview of the proof. We identify two key properties of the problem. The first property is that for all but at most two layers the cost structure is homogeneous. All costs $c_f(x_{i,j})$ are either zippers or i·zippers. Moreover, layers that only contain zippers (i·zippers) are consecutive. The layers $[n_2, \ldots, n]$ only contain zippers for some $n_2$. The layers $[1, \ldots, n_1]$ only contain i·zippers for some $n_1 < n_2$. There are at most two heterogeneous layers between these sequences.

*Example 5.* Consider Figure 3. We only show the forward cost $c_f(x_{i,j})$ for each variable-value pair in a gray rectangle. The homogeneous consecutive layers $[n_2 = 4, \ldots, n = 8]$ only contain zippers. The two heterogeneous consecutive layers $[2, 3]$ contain zippers and i·zippers. The homogeneous consecutive layers $[0, n_1 = 1]$ only contain i·zippers. □

The second property is that if we consider all cost sets at one layer then their lower(upper) bounds are at most distance two from each other. This is stated as the closeness property of the structure in Theorem 1. Section 3.4 proves the first property and Section 3.5 proves the second property. The rest of the proof of Theorem 1 uses induction on the number of layers, taking these properties into account. These proofs are in the Appendices. Appendix C.1 proves Theorem 1 for the sequence of layers that only contain cost sets that are zippers. Moreover, it imposes an additional property on the structure of zippers. Appendix C.2 proves Theorem 1 for the two heterogeneous layers. This is the most tedious part of the proof using enumeration of all possible distinct structures of the forward(backward) cost. This enumeration is feasible because of the properties of the cost structure in the first sequence. Appendix C.3 proves Theorem 1 for the last sequence that only contains i·zippers. We show that no new cost structures may appear in this sequence. Overall, we prove that there are a bounded number of cost structures at each layer.

### 3.4 Partitioning of layers

The proof of Theorem 1 is based on the following lemma that partitions variables $X_1, \ldots, X_n$ into three groups based on the structure of the forward costs (the backward costs are similar, but the partition may be different).

**Lemma 4.** *Consider a* SEQBIN$(N, X, C, B)$ *with monotone $B$ and arbitrary $C$. Let $X_t = j$ be the first variable in the reverse order of variables such that there exists a value $j$ and an interval $[a, b]$, $a < b$ such that $[a, b] \subseteq c_f(x_{t,j})$, i.e., for all $t' \in [t+1, n]$, $j \in D(X_{t'})$, there does **not** exist $[a', b']$, $a' < b'$, such that $[a', b'] \subseteq c_f(x_{t',j})$. Then for all $c_f(x_{s,j})$, $j \in D(X_s)$, $s \in [1, t-2]$, there exists an interval $[a_{s,j}, b_{s,j}]$ such that $[a_{s,j}, b_{s,j}] \subseteq c_f(x_{s,j})$.*

*Proof.* Consider the pair of variables $X_t$ and $X_{t-1}$. We recall that we consider variables in the reverse order form $n$ to 0. Let $v$ be the maximum value in the total order $\pi$ such that $v \in D(X_{t-1})$. By the monotonicity of $B$ and the fact that $B(X_{t-1}, X_t)$ is DC, we conclude that $(v, j) \in B$. Otherwise, if $(v, j) \notin B$, the value $j$ had to be pruned from $D(X_i)$ by enforcing DC on $B(X_{t-1}, X_t)$ as $v$ is the top value in the ordering in $D(X_{t-1})$. Therefore, there exists an interval $[c, d] \in \{[a, b], [a+1, b+1]\}$, $c < d$ such that $[c, d] \subseteq c_f(x_{t-1,v})$.

Consider the pair of variables $X_{t-1}$ and $X_{t-2}$. Due to monotonicity of $B$ we know that $(k, v) \in B$, $q \in D(X_{t-2})$ as $v$ is the top value in $\pi$ such that $v \in D(X_{t-1})$. Hence, $v$ is a support for all $k$ and $c_f(x_{t-2,k})$ must contain an interval as $c_f(x_{t-2,k}) = \bigcup_{w \in D(X_{t-1})} (c_f(x_{t-1,w}) \uplus c(j, w))$ and $[c, d] \subseteq c_f(x_{t-1,v})$, $v \in D(X_{t-1})$. Hence, there exists an interval $[c', d']$, $c' < d'$ such that $[c', d'] \subseteq c_f(x_{t-2,k})$ for all $k \in D(X_{t-2})$ including the top value in the ordering $\pi$, $k'$, such that $k' \in D(X_{t-2})$. We repeat the argument for layers $s$, $s \in [1, \ldots, t-3]$. □

**Corollary 1.** *Consider a* SEQBIN$(N, X, C, B)$ *with monotone $B$ and arbitrary $C$. Then there are three blocks of consecutive variables $[X_1, X_{n_1}] \cup [X_{n_1+1}, X_{n_2-1}] \cup [X_{n_2}, X_n]$ with $n_1 < n_2 \leq n_1 + 3$, i.e., the size of the partition $[X_{n_1+1}, X_{n_2-1}]$ is at most 2, and:*

**Zipper block.** *For all $i, j$, $i \in [n_2, n]$, $j \in D(X_i)$, there does* not *exist an interval $[a, b] \subseteq [1, \pi^{top}]$, $a < b$, such that $[a, b] \subseteq c_f(x_{i,j})$.*

**Zipper + i·Zipper block.** *There exist $i, j$, $i \in [n_1 + 1, n_2 - 1]$, $j \in D(X_i)$ and an interval $[a, b] \subseteq [1, \pi^{top}]$, $a < b$, such that $[a, b] \subseteq c_f(x_{i,j})$.*

**i·Zipper block.** *For all $i, j$, $i \in [1, n_1]$, $j \in D(X_i)$ there* exists *an interval $[a, b] \subseteq [1, \pi^{top}]$, $a < b$, such that $[a, b] \subseteq c_f(x_{i,j})$.*

*Example 6.* Consider Figure 3. The zipper block includes $[X_4, \ldots, X_8]$. The zipper + i·zipper block includes variables $X_2$ and $X_3$. The i·zipper block contains $X_1$. □

### 3.5 Closeness of costs

We show that if $B$ is a monotone constraint then the forward cost of the values of a variable cannot deviate too much from each other. Hence, we prove the closeness property of the cost structure in Theorem 1.

**Lemma 5.** *Consider a* SEQBIN$(N, X, C, B)$ *with monotone $B$ and arbitrary $C$. Consider a variable $X_i$, $i = [1, \ldots n]$. Then for any two values $j, k \in D(X_i)$, $j \prec k$, either $ub_f(x_{i,j}) \in [ub_f(x_{i,k}), ub_f(x_{i,k}) + 1]$ or $ub_f(x_{i,k}) \in [ub_f(x_{i,j}), ub_f(x_{i,j}) + 2]$.*

*Proof.* By induction on the distance from $n$. The base case is trivial, as $ub_f(x_{n,i}) = lb_f(x_{n,i}) = 1$ for all $i$. Suppose this holds for all $X_{t+1}, \ldots, X_n$. We show that it holds for $X_t$. Let $v$ be a value such that $ub_f(x_{t,k}) = ub_f(x_{t+1,v}) + c(k,v)$ and $w$ be a value such that $ub_f(x_{t,j}) = ub_f(x_{t+1,w}) + c(j,w)$.

*Property 1.* If $w = v$ or $ub_f(x_{t+1,v}) = ub_f(x_{t+1,w})$ then lemma holds. Proof: This follows from the assumption that all costs in $C$ are zero or one. Hence, $|ub_f(x_{t,j}) - ub_f(x_{t,k})| \le 1$.

*Property 2.* The tuple $(k, w) \in B$. Proof: This follows from monotonicity of $B$ and the assumption that $j \prec k$ and from $(j, w) \in B$.

*Property 3.* If $w \prec v$ then $(j, v) \in B$. Proof: This follows from monotonicity of $B$ and the assumptions $w \prec v$ and $(j, w) \in B$.

*Property 4.* If $(j, v) \in B$ then $w = v$. Proof: In this case the bipartite subgraph over four vertices $k, j, v, w$ is complete (Figure 4(a)). Hence, $v' = argmax_{w,v}(ub_f(x_{t+1,v}), ub_f(x_{t+1,w}))$ is a potential support for both $ub_f(x_{t,j})$ and $ub_f(x_{t,k})$ and $w$ and $v$ coincide.

From Properties 1–4 we know that we only have to prove Lemma in the following case: $v \prec w$, $v \ne w$, $ub_f(x_{t+1,v}) \ne ub_f(x_{t+1,w})$ and $(j, v) \notin B$.

By the induction hypothesis, there exist two cases: $ub_f(x_{t+1,v}) \in [ub_f(x_{t+1,w}), ub_f(x_{t+1,w}) + 1]$ or $ub_f(x_{t+1,w}) \in [ub_f(x_{t+1,v}), ub_f(x_{t+1,v}) + 2]$.

*Case 1.* We assume $ub_f(x_{t+1,v}) \in [ub_f(x_{t+1,w}), ub_f(x_{t+1,w}) + 1]$. As $ub_f(x_{t+1,v}) \ne ub_f(x_{t+1,w})$ we know that $ub_f(x_{t+1,v}) = ub_f(x_{t+1,w}) + 1$. We denote $p = ub_f(x_{t+1,w})$. Figure 4(b) shows this case. Note as costs of the edges are zero or one, $ub_f(x_{t,k}) \in \{p+1, p+2\}$. On the other hand, $ub_f(x_{t,j}) \in \{p, p+1\}$. Hence, $ub_f(x_{t,k}) \in [ub_f(x_{t,j}), ub_f(x_{t,j}) + 2]$ as required.

*Case 2.* We assume $ub_f(x_{t+1,w}) \in [ub_f(x_{t+1,v}), ub_f(x_{t+1,v}) + 2]$ (Figure 4 (c)) and since $ub_f(x_{t+1,v}) \ne ub_f(x_{t+1,w})$, $ub_f(x_{t+1,w}) > ub_f(x_{t+1,v})$. As $v \prec w$ the value $w$ is a support value for both $j$ and $k$. Hence, either $v = w$ or $ub_f(x_{t+1,v}) = ub_f(x_{t+1,w})$. This contradicts $v \ne w$ and $ub_f(x_{t+1,v}) \ne ub_f(x_{t+1,w})$. □

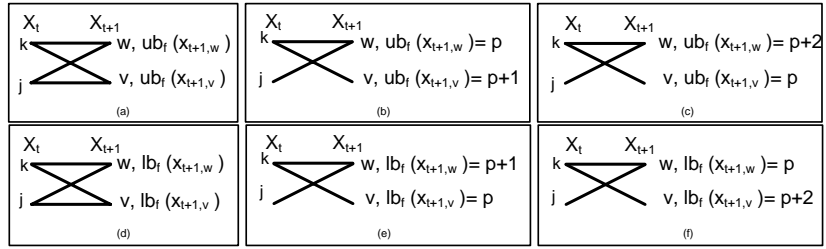

**Fig. 4.** Computation of the forward cost upper bound, $ub_f(x_{i,j})$, (a)–(c) and the forward cost lower bound, $lb_f(x_{i,j})$, (d)–(f). Note that we do not distinguish between 0 and 1 cost edges.

**Lemma 6.** *Consider a* SEQBIN$(N, X, C, B)$ *with monotone $B$ and arbitrary $C$. Then, either $lb_f(x_{i,j}) \in [lb_f(x_{i,k}), lb_f(x_{i,k}) + 2]$ or $lb_f(x_{i,k}) \in [lb_f(x_{i,j}), lb_f(x_{i,j}) + 1]$ for all variables $X_i$ and values $j \prec k$.*

*Proof.* Analogous to Lemma 5 (Figure 4, (d)–(f)). □

We omit the rest of the proof here due to space limitation (see Appendix B–C ). We only mention Appendix C.1, Lemma 15 that refines Theorem 1 for layers in the zipper block as we use this result in Section 4. Lemma 15 shows that at the $i$th layer in the zipper block, $i \in [n_1 + 3, n]$, there are at most 4 possible distinct sets $c_f(x_{i,j}), j \in D(X_i)$.

### 3.6 Total cost

**Lemma 7.** *Consider a* SEQBIN$(N, X, C, B)$ *constraint with monotone $B$ and arbitrary $C$. The set $c(x_{i,j}) = c_f(x_{i,j}) \uplus c_b(x_{i,j}) \uplus (-1)$, $j \in D(X_i)$, $i = 1, \ldots, n$ is either a zipper or an i·zipper set. For any i·zipper set $c(x_{i,j}) = [a \sim b - c \sim d]$ it holds $b - a \leq 4$ and $d - c \leq 4$. Moreover, $c(x_{i,j})$ can be computed in $O(1)$ time.*

*Proof.* It is sufficient to consider $c(x_{i,j}) = c_f(x_{i,j}) \uplus c_b(x_{i,j})$ as a shift by a constant does not change the structure of the set. As $c_f(x_{i,j})$ and $c_b(x_{i,j})$ satisfy Theorem 1, they are either zipper or i·zipper sets. We consider 3 cases.

*Case 1. Both $c_f(x_{i,j})$ and $c_b(x_{i,j})$ are zipper sets.* Consider zipper sets $c_f(x_{i,j}) = [a \sim b] = \{a, a+2, \ldots, b\}$ and $c_b(x_{i,j}) = [c \sim d] = \{c, c+2, \ldots, d\}$. Then $c(x_{i,j}) = \{a+c, a+2+c, \ldots, b+c, b+c+2, \ldots, \ldots b+d\} = [(a+c) \sim (b+d)]$.

*Case 2. Both $c_f(x_{i,j})$ and $c_b(x_{i,j})$, are i·zipper sets.* Consider $c_f(x_{i,j}) = [a \sim b - r \sim q]$ and $c_b(x_{i,j}) = [c \sim d - f \sim e]$. We consider the most general case where $a < b, r < q, c < d$ and $f < e$.

We perform the operation $\uplus$ in three steps, $c(x_{i,j}) = c^1 \cup c^2 \cup c^3$ where $c^1 = c_f(x_{i,j}) \uplus [d, f] = [a + d, q + f]$, $c^2 = c_f(x_{i,j}) \uplus [c \sim d] = [(a + c) \sim (b + d)] \cup [b + c, r + d] \cup [(r + c) \sim (b + q)]$, and $c^3 = c_f(x_{i,j}) \uplus [f \sim e] = [(a + f) \sim (b + e)] \cup [b + f, r + e] \cup [(r + f) \sim (e + q)]$. As $(b + c) \leq (b + d)$ and $(r + c) \leq (r + d)$ we get $c^2 = [(a + c) \sim (b + c) - (r + d) \sim (b + q)]$. Similarly, we get $c^3 = [(a + f) \sim (b + f) - (r + e) \sim (e + q)]$.

Finally, $c(x_{i,j}) = c^1 \cup c^2 \cup c^3 = [(a+c) \sim (\min((b+c), (a+d)) - \max((r+e), (q+f)) \sim (e+q)]$. Consider the value $\min((b+c), (a+d) - (a+c))$. If $b+c \leq a+d$ then we have $(b+c) - (a+c) = b - a \leq 4$. If $a+d \leq b+c$ then we have $(a+d) - (a+c) = d - c \leq 4$. Similarly, we prove the result $(e + q) - \max((r + e), (q + f)) \leq 4$. Hence, the statement of the lemma holds.

*Case 3. Exactly one of $\{c_f(x_{i,j}), c_b(x_{i,j})\}$ is a zipper set.* Similar to Case 2.

*Complexity.* In all three cases above, the proof is constructive and we give an analytic expression to compute $c(x_{i,j})$. Hence, this can be done in $O(1)$ time. □

*Example 7.* Suppose $c_f(x_{i,j}) = [2 \sim 6 - 8 \sim 12]$ and $c_b(x_{i,j}) = [10 \sim 16 - 20 \sim 22]$. Both $c_f(x_{i,j})$ and $c_b(x_{i,j})$ are i·zipper sets. Hence, to compute $c(x_{i,j}) = (c_f(x_{i,j}) \uplus c_b(x_{i,j}) \uplus (-1))$ we use the expression $[(2+10) \sim (\min((6+10), (2+16)) - \max((8+22), (12+20)) \sim (12+22)] \uplus (-1) = [11 \sim 15 - 31 \sim 33]$. □

## 4 Domain consistency algorithm

In this section we present SEQBINALGNEW, a domain consistency algorithm for SEQBIN($N, X, C, B$) with monotone $B$. It has the same structure as SEQBINALG:

**Phase 1** Remove all non $B$-coherent values in the domains of $X$.
**Phase 2** For all values in the domains of $X$, compute $c_f(x_{i,j})$ and $c_b(x_{i,j})$.
**Phase 3** Prune the domain of $N$ with respect to $c_f(x_{0,0^*})$.
**Phase 4** Prune the remaining $B$-coherent values.

The main complexity bottleneck is Phase 2 and Phase 4. If we do not put any restrictions on $B$ and $C$ then it takes $O(n^2 d^2)$ in total to compute these sets. We show that the complexity of SEQBINALGNEW decreases as we put restrictions on constraints $B$ and $C$. With respect to phase 3, we note that the cardinality of both $D(N)$ and $c_f(x_{0,0^*})$ is at most $n$, so their intersection can be computed in time $O(n)$.

### 4.1 Domain consistency algorithm in $O(nd^2)$ with monotone $B$

*Phase 2 of* SEQBINALGNEW. We exploit the structure of the costs established by Theorem 1 to improve PATHDP (Phase 2). We show that lines 4–5 and 8–9 can be done in $O(d)$ time if $B$ is monotone.

**Lemma 8.** *Consider a* SEQBIN($N, X, C, B$) *constraint such that $B$ is monotone. For all $j \in D(X_i)$, $i \in [1, \ldots, n]$, $c_f(x_{i,j}) = \bigcup_{v \in D(X_{i+1})} (c_f(x_{i+1,v}) \uplus c(j, v))$ can be computed in $O(d)$ time.*

*Proof.* We partition all supports $v$ into two groups based on the value of $c(j, v)$. The first group $S_0$ contains values such that $c(j, v) = 0$ and the second group $S_1$ contains values such that $c(j, v) = 1$. We find $c^1 = \bigcup_{v \in S_0} c_f(x_{i+1,v})$ and $c^2 = \bigcup_{v \in S_1} c_f(x_{i+1,v})$. Then we find $c_f(x_{i,j}) = c^1 \cup (c^2 \uplus 1)$. We prove the lemma for $c^1$ ($c^2$ is analogous.)

*Compute $c^1$.* We assume that $p$ is the smallest lower bound among the forward cost sets of the values in $S_0$ and $q + 2$ is the greatest upper bound: $p = \min_{v \in S_0} lb_f(x_{i+1,v})$ and $q + 2 = \max_{v \in S_0} ub_f(x_{i+1,v})$. We refer to $l \cdot zip$ of $c_f(x_{i+1,v})$ as $l \cdot zip(x_{i+1,v})$ to simplify notation (similarly, for the other two parts $i \cdot val$ and $r \cdot zip$). By Theorem 1 we know that $lb(l \cdot zip(x_{i+1,v})) \in [p, p+2]$, $ub(r \cdot zip(x_{i+1,v})) \in [q, q+2]$, $lb(i \cdot val(x_{i+1,v})) \in [p, p+6]$ and $ub(i \cdot val(x_{i+1,v})) \in [q-4, q+2]$. Hence, we compute the 20 *indicator values* $J_y^{l \cdot zip}(v)$, $y \in [p, p+2]$, $J_y^{r \cdot zip}(v)$, $y \in [q, q+2]$, $J_y^{i \cdot val_{lb}}(v)$, $y \in [p, p+6]$, and $J_y^{i \cdot val_{ub}}(v)$, $y \in [q-4, q+2]$, $v \in S_0$. For example, we define $J_y^{l \cdot zip}(v) = 1$, iff $lb(l \cdot zip(x_{i+1,v})) = y$ and $J_y^{l \cdot zip}(S_0) = \max_{v \in S_0} J_y^{l \cdot zip}(v)$, $y \in [p, p+2]$. Similarly, we compute the other 19 indicators. This can be done in $O(d)$ time with a linear scan over $c_f(x_{i+1,v})$, $v \in S_0$. Then we can compute $\bigcup_{v \in S_0} c_f(x_{i+1,v}) = [a^* \sim b^* - c^* \sim d^*]$ in 4 steps, each of which takes $O(1)$ time.

*Union of $i \cdot val$.* Theorem 1 shows that all $i \cdot val$ sets must overlap. Hence, the union of $i \cdot val(x_{i+1,j})$ forms an interval. We find the minimum value $y$, $y \in \{p, \ldots, p+6\}$ such that $J_y^{i \cdot val_{lb}}(S_0) = 1$. If such a value $y$ exists then we set $b^* = y$. Then we find the largest value $y' \in \{q-4, \ldots, q+2\}$ such that $J^{i \cdot val_{ub}}_{y'}(S_0) = 1$ and set $c^* = y'$. Note that if $y$ exists then $y'$ exists. If $y$ does not exist we know that all $c_f(x_{i+1,v})$, $v \in S_0$ are zipper sets and we set $b^* = c^* = \emptyset$.

*Union of $l \cdot zip$.* Suppose $b^* \neq \emptyset$. We find indicators $J^{l \cdot zip}_y(S_0)$, $y \in [p, p+2]$, that are set to one. Set $p'$ to the minimum among $[p, p+2]$, for which there exists $J^{l \cdot zip}_{p'}(S_0) = 1$. If $J^{l \cdot zip}_{p+1}(S_0) = 1$ and $J^{l \cdot zip}_p(S_0) = 1$ or $J^{l \cdot zip}_{p+2}(S_0) = 1$ or $b^* \in \{p, p+2\}$ then set $a^*$ and reset $b^*$, so that $a^* = b^* = \min(p+1, p')$ otherwise set $a^* = p'$ and leave $b^*$ unchanged. Union of $r \cdot zip$ is similar to union of $l \cdot zip$.

*Union of zippers.* Suppose $b^* = \emptyset$. Then we determine which of 4 distinct sets (Appendix C.1, Lemma 15) are present among $c^1_f(x_{i+1,v})$, $v \in S_0$. As there are at most 4 such that are zippers we can union them in $O(1)$ time and identify the values $a^*, b^*, c^*$ and $d^*$.

We can compute $c^1 \cup (c^2 \uplus 1)$ in $O(1)$. We omit the proof here due to space considerations (see Appendix D, Lemma 19).

*Complexity.* For each $j \in D(X_i)$, $i \in [1, \ldots, n]$, the forward cost set $c_f(x_{i,j})$, can be computed in $O(d)$. As we have $O(nd)$ such sets, the total time complexity is $O(nd^2)$. One way to reduce this complexity is to compute $c_f(x_{i,j})$ in $O(1)$. □

**Corollary 2.** *Phase 2 of the algorithm* SEQBINALGNEW *runs in $O(nd^2)$ time.*

*Phase 4 of* SEQBINALGNEW. We present the final phase of SEQBINALGNEW.

**Lemma 9.** *Consider a* SEQBIN$(N, X, C, B)$ *constraint such that $B$ is monotone. For each $i \in [1, \ldots, n]$, the **total** time complexity to compute $c(x_{i,j}) \cap D(N) \neq \emptyset$, $j \in D(X_i)$, is $O(d)$. The total time complexity of Phase 4 is $O(nd)$.*

*Proof. Preprocessing of $D(N)$.* We use a preprocessing step to compute cumulatively sums $s^{odd}_v$ and $s^{even}_v$ to collect information about the presence of odd and even values in $D(N)$. Hence, $s^{odd}_0 = 0$, $s^{odd}_{j+1} = s^{odd}_j + (j \in D(N) \land j \text{ is odd })$, $j \in [1, \ldots, \pi^{top}]$. Similarly, we compute $s^{even}_j$. This can be done in $O(d)$. Then the value $s^{odd}_{j_1} - s^{odd}_{j_2-1}$ shows how many odd values of $D(N)$ are in the interval $[j_2, j_1]$.

*Performing the check.* By Lemma 7 we know that $c(x_{i,j})$ is either zipper or i·zipper. If $c(x_{i,j})$ is an even zipper set $[a \sim b]$ we check if $s^{even}_b - s^{even}_{a-1} \neq 0$. If so the variable-value pair $X_i = j$ is supported. Similarly, if $c(x_{i,j})$ is an odd zipper set. Suppose $c(x_{i,j})$ is an i·zipper set $[a \sim b - c \sim d]$. Then, we can check separately whether each of three parts $[a \sim b] \cup [b - c] \cup [c \sim d]$ has an intersection with $D(N)$ using the cumulative sum values. Hence, the check can be done in $O(1)$ time. There are $O(d)$ sets $c(x_{i,j})$, $j \in D(X_i)$. Hence, the total time complexity of one layer is $O(d)$.

*Complexity.* The graph has $O(n)$ layers. So, the total time complexity is $O(nd)$. □

### 4.2 DC algorithm with monotone $B$ and row and column convex $C$

Finally, we show that if $C$ is row and column convex then SEQBINALGNEW runs in $O(nd)$ time. The only remaining bottleneck is Phase 2.

**Lemma 10.** *Consider a* SEQBIN$(N, X, C, B)$ *constraint such that $B$ is monotone and $C$ is row and column convex under that same ordering $\pi$ that gives monotonicity. The sets $c_f(x_{i,j})$ and $c_b(x_{i,j})$, $j \in D(X_i)$, $i \in [1, \ldots, n]$, can be computed in time $O(d)$.*

*Proof.* We give an algorithm to compute $c_f(x_{i,j})$. Computing $c_b(x_{i,j})$ is similar. Recall that in PATHDP (lines 4–5), $c_f(x_{i,j}) = \bigcup_{v \in D(X_{i+1}),(j,v) \in B} (c_f(x_{i+1,v}) \uplus c(j,v))$. Since $B$ is monotone, the set of $B$ supports of $X_i = j$, $Supports(x_{i,j}) = \{v | (j,v) \in B \wedge v \in D(X_{i+1})\}$, forms the *interval* $[a, v^t]$, $v^t \leq \pi^{top}$ for some $a$ such that $v^t$ is that maximum value in $D(X_{i+1})$.

As $C$ is row convex, the interval $[a, v^t]$ is partitioned into 3 subintervals $[a, v^t] = [a, b] \cup [b, c] \cup [c, v^t]$ such that $c(j, v) = 1$, $v \in [a, b] \cup [c, v^t]$ and $c(j, v) = 0$, $v \in [b, c]$ and we can write $c_f(x_{i,j}) = c^1 \cup c^2 \cup c^3$ where $c^1 = \bigcup_{v \in [a,b] \cap D(X_{i+1})} c_f(x_{i+1,v}) \uplus 1$, $c^2 = \bigcup_{v \in [b,c] \cap D(X_{i+1})} c_f(x_{i+1,v})$ and $c^3 = \bigcup_{v \in [c,d] \cap D(X_{i+1})} c_f(x_{i+1,v}) \uplus 1$. We exploit the fact that $c^1, c^2, c^3$ are computed over intervals to avoid recomputation of the indicator values for each $c(x_{i,j})$, as was necessary in Lemma 8. We do this with an $O(d)$ time preprocessing step that allows us to then compute each $c(x_{i,j})$ in $O(1)$. This reduces the complexity of lines 2–5 from $O(n^2 d^2)$ to $O(nd)$.

The preprocessing step consists in computing cumulative sums over the indicator values in an interval. For each indicator value $J_y^z$, $z \in \{l \cdot zip, r \cdot zip, i \cdot val\}$, we compute the array $cs_y^z(i)$, which counts the number of values in $[1, i]$ for which the indicator value is 1. For example, $cs_y^{l \cdot zip}(0) = 0$, $cs_y^{l \cdot zip}(v) = cs_y^{l \cdot zip}(v-1) + J_y^{l \cdot zip}(v)$, $y \in [p, p+2]$, $v \in D(X_{i+1})$. To compute the cumulative sums we do a linear scan over $c_f(x_{i+1,v})$, $v \in D(X_{i+1})$. Given these sums we can compute whether, for example, $lb(l \cdot zip(x_{i+1,v})) = y$, $v \in [a', b']$ in constant time by checking whether $cs_y^{l \cdot zip}(b') - cs_y^{l \cdot zip}(a'-1) > 0$.

The rest of the proof is identical to Lemma 8 (subsection 'Compute $c^1$'.). It takes $O(1)$ time to compute $c^1$ given the cumulative sums. We do this for $d$ sets $c_f(x_{i,j})$ and the preprocessing step takes $O(d)$, so the total time complexity is $O(d)$. □

**Corollary 3.** *Lemma 10 holds if the negation of $C$ is row and column convex under the same ordering $\pi$ that gives monotonicity.*

*Proof.* The only difference from the proof of Lemma 10 is that the interval $[a, v^t]$ of supports of each value $j \in D(X_i)$ is partitioned into $[a, v^t] = [a, b] \cup [b, c] \cup [c, v^t]$, such that $c(j, v) = 0$, $v \in [a, b] \cup [c, v^t]$ and $c(j, v) = 1$, $v \in [b, c]$. □

*Example 8.* Suppose $c_f(x_{i+1,v})$, $v \in [1, 2, 3]$ contain the following forward costs: $c_f(x_{i+1,1}) = [1 \sim 5 - 8 \sim 12]$, $c_f(x_{i+1,2}) = [3 \sim 5 - 6 \sim 10]$ and $c_f(x_{i+1,3}) = [2 \sim 6 - 8 \sim 10]$. The min value $p$ is 1 and the max value $q+2$ is 12. First we compute cumulative sums. The table below shows the non-zero vectors of cumulative sums.

| | $v$ | | | $v$ | | | $v$ | | | $v$ |
|---|---|---|---|---|---|---|---|---|---|---|
| Cumulative sums | 0 1 2 3 | Cumulative sums | 0 1 2 3 | Cumulative sums | 0 1 2 3 | Cumulative sums | 0 1 2 3 |
| $cs_1^{l \cdot zip}(v)$ | [0 1 1 1] | $cs_{10}^{r \cdot zip}(v)$ | [0 0 1 2] | $cs_5^{i \cdot val_{lb}}(v)$ | [0 1 2 2] | $cs_6^{i \cdot val_{ub}}(v)$ | [0 0 1 1] |
| $cs_2^{l \cdot zip}(v)$ | [0 0 0 1] | $cs_{12}^{r \cdot zip}(v)$ | [0 1 1 1] | $cs_6^{i \cdot val_{lb}}(v)$ | [0 0 0 1] | $cs_8^{i \cdot val_{ub}}(v)$ | [0 1 1 2] |
| $cs_3^{l \cdot zip}(v)$ | [0 0 1 1] | | | | | | |

Suppose that the values $1, 2$ and $3$ are supports for $c_f(x_{i,1}) = [a^* \sim b^* - c^* \sim d^*]$. Using Lemma 10, we find $\cup_{j=[1,3]} i \cdot val(x_{i+1,j}) = [5-8] \cup [5-6] \cup [6-8] = [5-8]$. So $b^* = 5$ and $c^* = 8$ Then, we check if there exists $lb(l \cdot zip(x_{i+1,j})) = y, y \in \{1, 2, 3\}$ using cumulative sums. For $y \in \{1, 2, 3\}$ we get that $cs_y^{l \cdot zip}(3) - cs_y^{l \cdot zip}(0) > 0$. So we set $a^*$ and reset $b^*$ so that $a^* = b^* = \min(2, 1) = 1$. Finally, we check if there exists $ub(r \cdot zip(x_{i+1,j})) = y, y \in \{10, 11, 12\}$. For $y \in \{10, 12\}$ we get that $cs_y^{r \cdot zip}(3) - cs_y^{l \cdot zip}(0) > 0$. Moreover, the value $q + 1 = 11$ does not occur among $ub(r \cdot zip(x_{i+1,j}))$. Hence, we set $d^* = 12$. This gives $c_f(x_{i,1}) = [1 - 8 \sim 12]$. □

**Corollary 4.** *The filtering algorithm* SEQBINALGNEW *enforces domain consistency on* CHANGE *and* SMOOTH *in* $O(nd)$ *time.*

*Proof.* CHANGE is SEQBIN($N, X, C \in \{=, \neq, <, \leq, >, \geq\}$, TRUE). This satisfies Lemma 10 as $\{=, <, \leq, >, \geq\}$ are row/column convex as is the negation of $\{\neq\}$. SMOOTH is SEQBIN($N, X, C$ is $\{|X_i - X_{i+1}| > cst\}$, TRUE), $cst \in N$, is the negation of row/column convex constraint $\{|X_i - X_{i+1}| \leq cst\}$. □

**Corollary 5.** *The filtering algorithm* SEQBINALGNEW *enforces domain consistency on* INCREASINGNVALUE *in* $O(nd)$ *time.*

*Proof.* INCREASINGNVALUE($X, N$) is SEQBIN($X, N, \leq, =$) [1]. This version of the SEQBIN constraint is counting-continuous and therefore $c(x_{i,j})$ is an interval. Hence, all costs $c_f, c_b$ are intervals. Moreover, $=$ is row and column convex, so SEQBINALGNEW reduces to SEQBINALG and enforces GAC in $O(nd)$. □

Finally, we note that we can slightly generalize SEQBIN so that it does not require the same $B$ and $C$ for every pair of variables as the proof of Theorem 1 does not rely on the property that $B$ and $C$ are the same for each pair of consecutive variables.

## 5 Conclusions

The SEQBIN meta-constraint subsumes a number of important global constraints like CHANGE, SMOOTH and INCREASINGNVALUE. We have shown that the filtering algorithm for SEQBIN proposed in [1] has two drawbacks even under strong restrictions: it does not detect bounds disentailment and it is not idempotent. We identified the cause for these problems, and proposed a new propagator that overcomes both issues. Our algorithm is based on a connection to the problem of finding a path of a given cost in a restricted $n$-partite graph. Our propagator enforces domain consistency in $O(nd^2)$ and, for special cases of SEQBIN that include CHANGE, SMOOTH, and INCREASINGNVALUE, in $O(nd)$ time.

## A  Revisiting SEQBIN.

**Lemma 11.** *The filtering algorithm* SEQBINALG *for* SEQBIN$(N, X, C, B)$ *with non-monotone B is not idempotent. There is a family of problems where* SEQBINALG *takes* $O(n)$ *iterations to reach the fixpoint.*

*Proof.* Figure 5(a) shows the graph representation of SEQBIN$(N, [X_1, \ldots, X_9], C, B)$, $D(N) = [2, 4, 6, 8]$.

*Iteration 1.* The minimum cost assignment is $[0, 1, 0, 1, 0, 1, 0, 1, 0, 4]$ with cost 1. Moreover, this assignment is the only support for $X_8 = 0$. As $1 \notin D(N)$, $X_8 = 0$ is pruned. None of the other value is removed. The first iteration is finished. Figure 5(b) shows new domains.

*Iteration 2.* The minimum cost assignment is $[0, 1, 0, 1, 0, 1, 0, 1, 2, 4]$ with cost 3. Moreover, this assignment is the only support for $X_6 = 0$ and $X_7 = 1$. As $3 \notin D(N)$, these values are pruned. Figure 5(c) shows new domains.

*Iteration 3.* The minimum cost assignment is $[0, 1, 0, 1, 0, 1, 2, 2, 2, 4]$ with cost 5. Moreover, this assignment is the only support for $X_4 = 0$ and $X_5 = 1$. As $5 \notin D(N)$, these values are pruned.

The reason for this behavior is that possible costs are $[1, 3, 5, 7, 8]$, but only assignments with cost 8 are solutions. As SEQBINALG only computes lower and upper bounds on the cost it prunes values that are supported by an assignment with cost 1 first, then it prunes values that are supported by an assignment with cost 3 and so on.

Note that this example embeds a pattern of two variables, e.g. $X_4, X_5$ and $X_6, X_7$. We can extend the example with an unbounded number of patterns, $p = O(n)$, on two variables and adjust $D(N)$ appropriately. Hence, SEQBINALG takes $p/2$ iterations to reach the fixpoint. □

## B  Other properties of closeness of costs

The following lemma shows that there exists only one way to obtain forward cost sets for two variable-value pairs $X_i = j$ and $X_i = k$ such that the distance between their upper bounds is two.

**Lemma 12.** *Consider a* SEQBIN$(N, X, C, B)$ *with monotone B and arbitrary C. Consider two values* $j, k \in D(X_i)$, $j \prec k$. *Suppose* $ub_f(x_{i,j}) = p$ *and* $ub_f(x_{i,k}) = p + 2$. *The following holds:*

1. $\forall u \in D(X_{i+1}), (j, u) \in B$ *we have that* $ub_f(x_{i+1,u}) = p$
2. $\exists v, w \in D(X_{i+1})$, $v \prec w$, $(j, v) \notin B$ *and* $(j, w) \in B$ *such that* $ub_f(x_{i+1,v}) = p + 1$, $ub_f(x_{i+1,w}) = p$, $c(k, v) = 1$ *and* $c(j, w) = 0$.

*Proof.* Figure 6 (a) illustrates the lemma.

Let $v, w$ be values in $D(X_{i+1})$ such that we obtain $ub_f(x_{i,k})$ from $ub_f(x_{i+1,v})$ and $ub_f(x_{i,j})$ from $ub_f(x_{i+1,w})$.

First, we observe that $ub_f(x_{i+1,u}) + c(j, u) \leq p$, otherwise $ub_f(x_{i,j}) > p$. As the cost value is 0 or 1 this implies $ub_f(x_{i+1,u}) \leq p$.

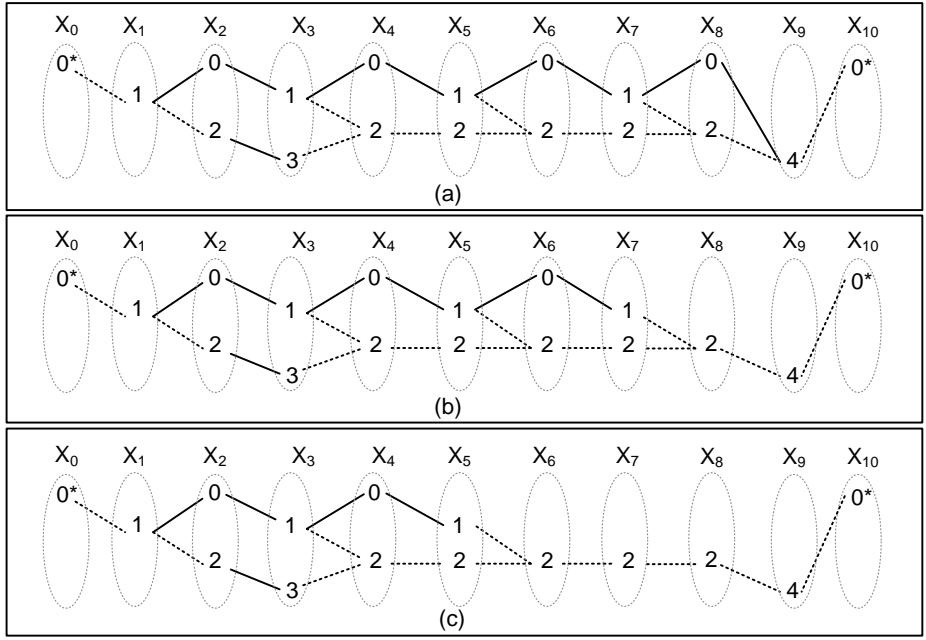

**Fig. 5.** A 11-partite graph that corresponds to the SEQBIN constraint from the proof of Lemma 11. Dashed edges have cost one and solid edges have cost zero. (a) shows initial costs; (b) shows costs after $X_3 = 0$ is pruned.

Second, if $(j, v) \in B$ then, as $(j, w) \in B$ and costs are 0 or 1, $|ub_f(x_{i,k}) - ub_f(x_{i,j})| \leq 1$, a contradition. Hence, $(j, v) \notin B$.

Third, we can only obtain $ub_f(x_{i,j}) = p$ and $ub_f(x_{i,k}) = p + 2$ if $v \prec w$ and $ub_f(x_{i+1,w}) < ub_f(x_{i+1,v})$. Otherwise, if $w \prec v$, then from $(j, w) \in B$ and the monotonicity of $B$, we get $(j, v) \in B$, which we showed above is contradictory. If $v \prec w$ and $ub_f(x_{i+1,w}) \geq ub_f(x_{i+1,v})$ then, because $v \prec w$ and $B$ is monotone, the value $ub_f(x_{i+1,w})$ or $ub_f(x_{i+1,w}) + 1$ is in both sets $c_f(x_{i,j})$ and $c_f(x_{i,k})$. Hence, $|ub_f(x_{i,k}) - ub_f(x_{i,j})| \leq 1$, a contradiction.

Therefore, $ub_f(x_{i,j}) = p$ and $ub_f(x_{i,k}) = p + 2$ must be obtained from two support values $v, w$ such that $v \prec w$ and $ub_f(x_{i+1,w}) < ub_f(x_{i+1,v})$ and $(j, v) \notin B$. By Lemma 5, if $v \prec w$ and $ub_f(x_{i+1,v}) > ub_f(x_{i+1,w})$ then $ub_f(x_{i+1,v}) = ub_f(x_{i+1,w}) + 1$. As $ub_f(x_{i,j}) = p$ and $(j, v) \notin B$ then there are two cases.

Suppose $ub_f(x_{i+1,w}) = p$ and $c(j, w) = 0$. As $ub_f(x_{i,k}) = p + 2$ then $ub_f(x_{i+1,v}) = p + 1$.

Suppose $ub_f(x_{i+1,w}) = p - 1$ and $c(j, w) = 1$. As $ub_f(x_{i,k}) = p + 2$ then $ub_f(x_{i+1,v})$ must be equal to $p+1$. and $c(j, w) = 1$. As $ub_f(x_{i+1,v}) - ub_f(x_{i+1,w}) = 2$ then $w \prec v$. This contradicts the fact that $v \prec w$.

Hence the only possible upper bound costs for support values are $ub_f(x_{i+1,w}) = p$ and $ub_f(x_{i+1,v}) = p + 1$.

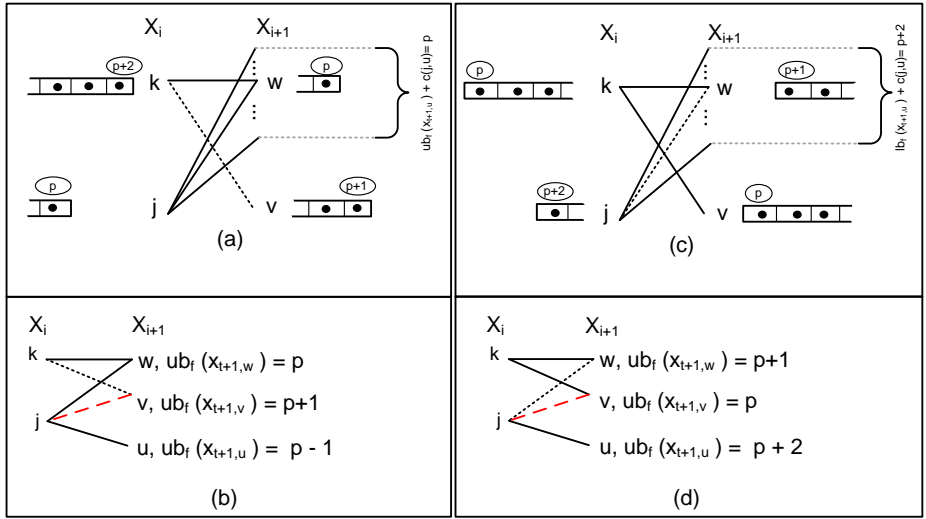

**Fig. 6.** Computation of the upper bound on the forward cost for $j, k \in D(X_i)$, $j \prec k$ (a) $ub_f(x_{i,j}) = p$ and $ub_f(x_{i,k}) = p + 2$; (b) demonstrates that $\nexists u \in D(X_{i+1}), (j, u) \in B$ such that $ub_f(x_{i+1,u}) = p - 1$ in case (a); (c) $lb_f(x_{i,j}) = p + 2$ and $lb_f(x_{i,k}) = p$; (d) demonstrates that $\nexists u \in D(X_{i+1}), (j, u) \in B$ such that $lb_f(x_{i+1,u}) = p + 2$ in case (c);

Finally, we prove that $ub_f(x_{i+1,u}) = p$ for all supports $u$ of $j$ in $X_{i+1}$. First, by lemma 5, and $ub_f(x_{i+1,v}) \geq p + 1$, we get $ub_f(x_{i+1,u}) \geq p - 1$. Suppose there exists $u \in D(X_{i+1}), (j, u) \in B$ such that $ub_f(x_{i+1,u}) = p - 1$. We know that there exists a support value $v$, $ub_f(x_{i+1,v}) = p + 1$. Hence, $u \prec v$ by Lemma 5. As $(j, u) \in B$ then, by monotonicity $(j, v) \in B$ which contradicts the fact that $(j, v) \notin B$. Figure 6 (b) illustrates this situation. □

**Lemma 13.** *Consider a* SEQBIN$(N, X, C, B)$ *with monotone $B$ and arbitrary $C$. Consider two values $j, k \in D(X_i)$, $j \prec k$ such that $lb_f(x_{i,j}) = p + 2$ and $lb_f(x_{i,k}) = p$. The following holds:*

- $\forall u \in D(X_{i+1}), (j, u) \in B$ *we have* $ub_f(x_{i+1,u}) = p + 1$.
- $\exists v, w \in D(X_{i+1}), v \prec w, (j, v) \notin B$ *and* $(j, w) \in B$ *such that* $lb_f(x_{i+1,v}) = p$, $lb_f(x_{i+1,w}) = p + 1$ *and* $c(k, v) = 0$ *and* $c(j, w) = 1$.

*Proof.* The proof is analogous to Lemma 12 (Figure 6, (c) and (d)). □

## C  The inductive part of Theorem 1's proof

**Definition 5.** *An i·zipper $[a \sim b - c \sim d]$ is* non degenerated *if either $a < b$ or $c < d$.*

Note that an interval is a degenerated i·zipper set.

*Example 9.* The set $c_f(x_{1,3}) = [1 \sim 5 - 8]$ is non degenerated.

### C.1 Zipper block

We consider the first block of variables. We prove Theorem 1 for this block together with additional restrictions of a structure of cost.

**Lemma 14.** *Consider a* SEQBIN$(N, X, C, B)$ *constraint with monotone $B$ and arbitrary $C$. Suppose $c_f(x_{t,j})$, $j \in D(X_t)$, $t \in [n_2, n]$ does not contain an interval. Then for any $c_f(x_{t,j})$, $t \in [n_2, n]$, $j \in D(X_t)$ the following holds:*

- **Uniqueness and Overlapping.** $c_f(x_{t,j})$ *is a zipper set.*
- **Cost Structure (upper bounds).** *If there exist two values $j, k \in D(X_t)$ such that $ub_f(x_{t,k}) - ub_f(x_{t,j}) = 2$ and $ub_f(x_{t,k})$ is even(odd) then*
  1. $j \prec k$
  2. *if there exists $g$ such that $ub_f(x_{t,g})$ is odd (even) then $ub_f(x_{t,j}) < ub_f(x_{t,g}) < ub_f(x_{t,k})$.*
  3. $lb_f(x_{t,j}) = lb_f(x_{t,k})$
  4. *there do not exist two values $g, h \in D(X_t)$ such that $lb_f(x_{t,h}) - lb_f(x_{t,g}) = 2$ and $lb_f(x_{t,h})$ is even(odd).*
- **Cost Structure (lower bounds).** *If there exist two values $j, k \in D(X_t)$ such that $lb_f(x_{t,j}) - lb_f(x_{t,k}) = 2$ and $lb_f(x_{t,k})$ is even(odd).*
  1. $j \prec k$
  2. *if there exists $g$ such that $lb_f(x_{t,g})$ is odd (even) then $lb_f(x_{t,j}) < lb_f(x_{t,g}) < lb_f(x_{t,k})$.*
  3. $ub_f(x_{t,j}) = ub_f(x_{t,k})$
  4. *there do no exist two values $g, h \in D(X_t)$ such that $ub_f(x_{t,h}) - ub_f(x_{t,g}) = 2$ is even(odd).*

*Proof.* By induction on the distance from $n$. The base case is trivial, as $ub_f(x_{n,i}) = lb_f(x_{n,i}) = 1$ for all $i$. Suppose this holds for all $X_{t+1}, \ldots, X_n$. We show that it holds for $X_t$.

*Uniqueness and Overlapping.* We prove by contradiction. Suppose that $c_f(x_{t,j}) = [p, \ldots r, r + 3, q]$ so that there exists a hole of size 2. By induction hypothesis, $c_f(x_{t+1,j})$, $j \in D(X_{t+1})$ are zipper sets. Hence, $r$ and $r + 3$ must come from distinct support values $v$ and $w$, respectively. As $r \in c_f(x_{t,j})$ and $r + 2 \notin c_f(x_{t,j})$, $ub_f(x_{t+1,v}) + c(j, v) \leq r$. Hence, $lb_f(x_{t+1,v}) \leq ub_f(x_{t+1,v}) \leq r$.

As $r + 3 \in c_f(x_{t,j})$ and $r + 1 \notin c_f(x_{t,j})$, $lb_f(x_{t+1,w}) + c(j, w) \geq r + 3$. Hence, $ub_f(x_{t+1,w}) \geq lb_f(x_{t+1,w}) \geq r + 2$. Lemma 5, $v \prec w$ and $ub_f(x_{t+1,w}) = lb_f(x_{t+1,w}) = r + 2$. Putting it all together, $lb_f(x_{t+1,v}) \leq r$, $v \prec w$ and $lb_f(x_{t+1,w}) = r + 2$, we violate Lemma 6.

We only prove the properties of cost structure for the upper bounds, as the proofs for the lower bounds are symmetric.

*Cost Structure(upper bounds). Property 1.* If $ub_f(x_{t,k}) - ub_f(x_{t,j}) = 2$ then, by Lemma 5, $j \prec k$.

*Cost Structure(upper bounds). Property 2.* W.l.o.g. we assume that $ub_f(x_{t,k})$ is even. Consider $g$ such that $ub_f(x_{t,g})$ is odd.

Note that $ub_f(x_{t,g})$ can not be smaller than $ub_f(x_{t,j})$ as $ub_f(x_{t,k}) - ub_f(x_{t,g}) > 2$ in this case and Lemma 5 is violated. On the other hand, $ub_f(x_{t,g})$ can not be larger than

$ub_f(x_{t,k})$, as $ub_f(x_{t,g}) - ub_f(x_{t,j}) > 2$ and Lemma 5 is violated. Hence, $ub_f(x_{t,k}) < ub_f(x_{t,g}) < ub_f(x_{t,j})$.

*Cost Structure(upper bounds). Property 3.* We denote $lb_f(x_{t,k}) = p$. W.l.o.g. we assume that $p$ is even. Hence, $ub_f(x_{t,k})$ is even.

We prove by contradiction. If $lb_f(x_{t,j}) \neq lb_f(x_{t,k})$ then, by Lemma 6 and by the induction hypothesis $c_f(x_{t,j})$ and $c_f(x_{t,k})$ are zipper sets, hence, we get that $lb_f(x_{t,j}) = lb_f(x_{t,k}) + 2$.

By Lemma 13 we know that for all supports $u \in D(X_{t+1})$, $(j, u) \in B$ we have that $lb_f(x_{t+1,u}) = p + 1$. Hence, $lb_f(x_{t+1,u})$ is odd. As $c_f(x_{t+1,u})$ is a zipper set by the induction hypothesis, all supports of the forward cost of $j$ are odd zipper sets.

Consider the upper bounds on the forward cost. By assumption, $ub_f(x_{t,k}) - ub_f(x_{t,j}) = 2$. We denote $ub_f(x_{t,j}) = p'$. Note that $p'$ is even as $ub_f(x_{t,k})$ is even. By Lemma 12, we know that $j$ is only supported by values $u \in D(X_{t+1})$, $(j, u) \in B$ for which $ub_f(x_{t+1,u}) = p'$. As $c_f(x_{t+1,u})$ is a zipper set by the induction hypothesis, all supports of the forward cost of $j$ are even zipper sets, a contradiction.

*Cost Structure(upper bounds). Property 4.* We prove by contradiction for even $ub_f(x_{t,k})$ and $ub_f(x_{t,j})$. The odd case is symmetric. We denote $lb_f(x_{t,h}) = p$ and $ub_f(x_{t,k}) = q$.

Suppose there exist values $g, h \in D(X_t)$ such that $lb_f(x_{t,g}) - lb_f(x_{t,h}) = 2$ and $lb_f(x_{t,h})$ is even. Hence, $p$ and $q$ are even. By Lemmas 5–6 we also know that $j \prec k$ and $g \prec h$.

As $ub_f(x_{t,j}) \neq ub_f(x_{t,k})$ then $lb_f(x_{t,k}) = lb_f(x_{t,j})$ by Property 3 for the upper bounds. Hence, there are two options: $lb_f(x_{t,j}) = lb_f(x_{t,k}) = lb_f(x_{t,h}) = p$ or $lb_f(x_{t,j}) = lb_f(x_{t,k}) = lb_f(x_{t,g}) = p + 2$. In the first case, we have $lb_f(x_{t,g}) - lb_f(x_{t,k}) = 2$. In the second case, we have $lb_f(x_{t,k}) - lb_f(x_{t,h}) = 2$. Hence in both case, by Property 3 for the lower bounds, $ub_f(x_{t,j}) = ub_f(x_{t,g}) = ub_f(x_{t,h})$. This leads to a contradiction. □

We will show that there exist a bounded number of distinct sets $c_f(x_{i,j})$, $j \in D(X_i)$, $i = 1, \ldots, n$. Lemma 14 shows that all of them are zipper sets, however, these zipper sets might have different lower and upper bound that satisfy Lemma 14.

Lemma 15 enumerates of all distinct sets $c_f(x_{i,j})$, $j \in D(X_i)$. We know that each of $c_f(x_{i,j})$ is a zipper set ( and/or an i·zipper set in the next layers). However, these zippers are subsets of the interval $[p, q]$, $p = \min(lb_f(x_{i,j}))$ and $q = \max(lb_f(x_{i,j}))$, $j \in D(X_i)$. Hence to describe all possible $c_f(x_{i,j})$, in addition to the two possible structures, we need to take into account two differences: $lb_f(c_f(x_{i,j})) - p$ and $q - ub_f(c_f(x_{i,j}))$. As we use enumeration of all possible $c_f(x_{i,j})$ for other layers we define the notion of a *framed-structure* as one of the two possible structures, a zipper or an i·zipper, together with two differences $lb_f(c_f(x_{i,j})) - p$ and $q - ub_f(c_f(x_{i,j}))$. For example, $[p-1 \sim q-1]$ and $[p+2 \sim q+2]$ have the same structure, a zipper set, but there are two distinct *framed-structure* as they have different lower and upper bounds.

**Lemma 15.** *Consider a* SEQBIN$(N, X, C, B)$ *constraint with monotone $B$ and arbitrary $C$. Suppose $c_f(x_{t,j})$, $j \in D(X_t)$, $t \in [t', n]$ do not contain intervals. Let $p = \min_{j \in D(X_t)}\{lb_f(x_{t,j})\}$ and $q + 1 = \max_{j \in D(X_t)}\{ub_f(x_{t,j})\}$.*

1. Suppose there exist two values $j, k \in D(X_t)$ such that $ub_f(x_{t,k}) - ub_f(x_{t,j}) = 2$ and $ub_f(x_{t,k})$ is even(odd) and there exist two values $g, h \in D(X_t)$ such that $lb_f(x_{t,g}) - lb_f(x_{t,h}) = 2$ and $lb_f(x_{t,g})$ is odd(even). Then $\{j, g\} \prec \{k, h\}$. Moreover, $lb_f(x_{t,h}) < lb_f(x_{t,j}) = lb_f(x_{t,k}) < lb_f(x_{t,g})$ and $ub_f(x_{t,j}) < ub_f(x_{t,g}) = ub_f(x_{t,h}) < ub_f(x_{t,k})$.
2. $c_f(x_{t,j})$ is one of four framed-structure which are zipper sets together with their lower and upper bounds in Figure 7(a). We label 4 possible framed-structures of $c_f(x_{t,j})$ as $ST1, ST2, ST3$ and $ST4$.

*Proof. Property 1.* We only need to show that $g \prec k$ and $j \prec h$ as the rest follows from Lemmas 5–6.

We denote $lb_f(x_{t,h}) = p$ and $ub_f(x_{t,h}) = q$. Hence, $lb_f(x_{t,g}) = p + 2$ and $ub_f(x_{t,g}) = q$. As $ub_f(x_{t,k})$ and $ub_f(x_{t,j})$ are even and $ub_f(x_{t,j}) - ub_f(x_{t,k}) = 2$ then, by Lemma 14, Property 2 for the upper bounds we know that each odd zipper set must start at $p + 1$. Hence, $lb_f(x_{t,j}) = lb_f(x_{t,k}) = p + 1$. Therefore, taking into account that $ub_f(x_{t,h}) = q$, the only choices for the upper bounds of $k$ and $j$ are $ub_f(x_{t,k}) = q + 1$ and $ub_f(x_{t,j}) = q - 1$. Otherwise, we would violate Lemma 5. Hence $lb_f(x_{t,h}) = p < lb_f(x_{t,j}) = lb_f(x_{t,k}) = p + 1 < lb_f(x_{t,g}) = p + 2$ and $ub_f(x_{t,j}) = q - 1 < ub_f(x_{t,g}) = ub_f(x_{t,h}) = q < ub_f(x_{t,k}) = q + 1$. Figure 7(a) shows the structure of $c_f(x_{t,j}), c_f(x_{t,k}), c_f(x_{t,h})$ and $c_f(x_{t,g})$.

By an exhaustive search over all possible support values $v, w \in D(X_{t+1})$ that satisfy Lemmas 6–5, we find that the only possible structure of support values for each pair $j \prec k$ and $g \prec h$ is the following: $c_f(x_{t+1,v}) = [p \sim q]$ and $c_f(x_{t+1,w}) = [p+1 \sim q-1]$. Figure C.1 illustrate the support values $v, w$.

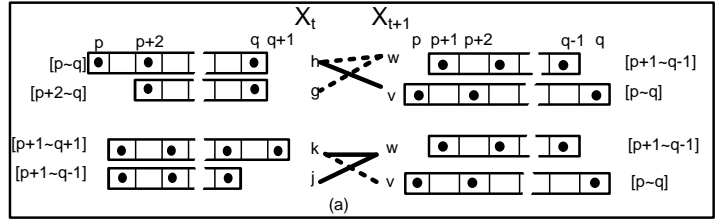

If $k \prec g$ and $(v, k) \in B$ then $(h, v) \in B$ and the $c_f(x_{t+1,v}) = [p \sim q]$ or $[p + 1 \sim q + 1]$ is included in $c_f(x_{t,g})$. Hence, $ub_f(x_{t,g}) \geq q$. This leads to a contradiction. The case $h \prec j$ is symmetric.

*Property 2.* The first property is immediate from Lemma 14 and Property 1. Figure 7(a) illustrates all possible framed-structures. □

*Example 10.* Consider the variable $X_4$ in Example 4. The set $c_f(x_{4,3}) = [1 \sim 5]$ is the framed-structure $ST2$ and $c_f(x_{4,2}) = [3 \sim 5]$ is the framed-structure $ST4$. □

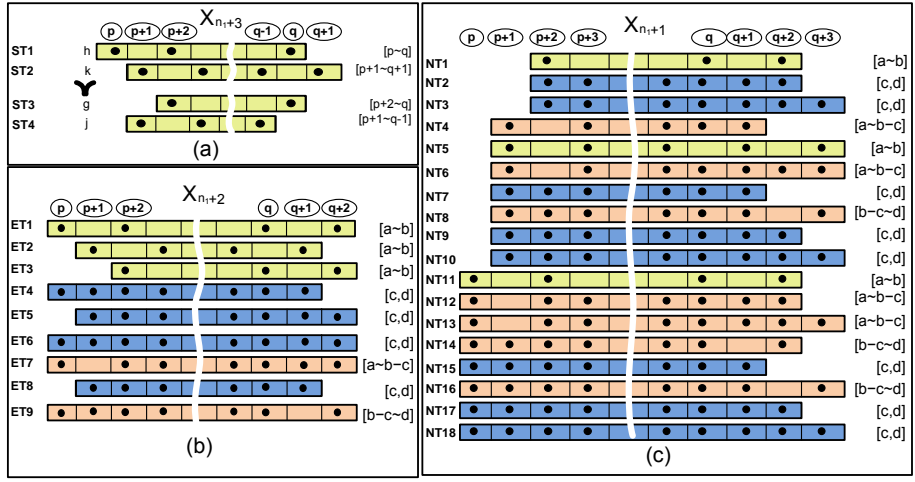

**Fig. 7.** Intervals are highlighted in blue, zipper sets are in yellow and an i·zipper sets are in orange. All possible framed-structures of (a) $c_f(x_{n_1+2}, j)$, $j \in D(X_{n_1+3})$; (b) $c_f(x_{n_1+2}, j)$, $j \in D(X_{n_1+2})$ and (c) $c_f(x_{n_1+1}, j)$, $j \in D(X_{n_1+1})$ when $ST1, ST2, ST3, ST4$ are present among $c_f(x_{n_2}, j)$;

### C.2 Zipper + i·Zipper block.

We consider the second group of variables. Lemma 4 shows that this group contains at most two variables. We assume here that it contains exactly two variables, as this is the most general case. Next we refine the structure of $c_f(x_i, j)$, $i \in [n_1 + 1, n_1 + 2]$ and prove Theorem 1.

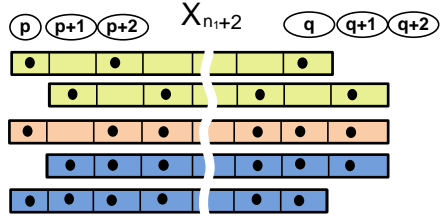

**Fig. 8.** Intervals are highlighted in blue, zipper sets are in yellow and an i·zipper sets are in orange. All possible framed-structures of $c_f(x_{n_1+2}, j)$, $j \in D(X_{n_1+2})$ when only $ST1, ST3, ST4$ are present among $c_f(x_{n_1+3}, j)$;

**Lemma 16.** *Consider a* SEQBIN$(N, X, C, B)$ *constraint with monotone $B$ and arbitrary $C$. Suppose $c_f(x_{s,j})$, $j \in D(X_s)$, $s \in [n_1 + 1, n_1 + 2]$ are zipper sets. Suppose there exist $j, j' \in D(X_t)$, $t = n_1 + 2$, and $k, k' \in D(X_{t-1})$ such that $c_f(x_{t,j})$ and*

$c_f(x_{t-1,k})$ are zipper sets and $c_f(x_{t,j'})$ and $c_f(x_{t-1,k'})$ are i·zipper sets. Then the following holds for $X_{t-1}$ and $X_t$:

- **Uniqueness.** $c_f(x_{t,j})$ is either a zipper set or an i·zipper set $[a \sim b - c \sim d]$ where either $a = b$ and/or $c = d$.
- **Overlapping.** If there exist two values $j, k \in D(X_t)$ such that $c_f(x_{t,j})$ and $c_f(x_{t,k})$ are two non-degenerated i·zipper sets $[a \sim b-c]$ and $[b' - c' \sim d]$(or $[d \sim b' - c']$), respectively, then $[b, c] \cap [b', c'] \neq \emptyset$.
- **Cost Structure.**
  1. If there exist two values $j, k \in D(X_t)$ such that $c_f(x_{t,j})$ and $c_f(x_{t,k})$ are two non-degenerated i·zipper sets $[a \sim b - c]$ and $[b' - c' \sim d]$ then $a = b'$ and $c = d$.
  2. If there exists a value $j \in D(X_t)$ such that $c_f(x_{t,j})$ are non-degenerated i·zipper sets $[a \sim b - c]$ ($[b' - c' \sim d]$) then $b = a + 2$ ($d = c' + 2$).

*Proof.* First, we prove the lemma for $X_t$. Lemma 15, Property 2, defines all possible structures of $c_f(x_{t+1,v})$, $v \in D(X_{t+1})$.

We perform complete enumeration of all possible structures of $c_f(x_{t,j})$, $j \in D(X_t)$ taking into account that $\{j, g\} \prec \{k, h\}$. Figure 7(b) shows the result of the enumeration if $ST1 - ST4$ set framed-structures are present among $c_f(x_{t+1,j})$, $j \in D(X_{t+1})$. We label all framed-structures as $ET1 - ET9$. We also show the type of each framed-structure. To keep presentation clear we omit actual values for the lower bound and upper bounds and use $[b, c]$ to denote an interval, $[a \sim b]$ to denote a zipper and $[a \sim b - c]$ or $[b - c \sim d]$ to denote an i·zipper. Note that if one of the framed-structures $ST1 - ST4$ is missing then we get a subset of framed-structures of $ET1 - ET9$. For example, Figure 8 shows the result of the enumeration if $ST1, ST3, ST4$ are present among $c_f(x_{t+1,j})$, $j \in D(X_{t+1})$. Hence, we can assume that all 4 framed-structures are presented among $c_f(x_{t+1,j})$, $j \in D(X_{t+1})$.

From the framed-structures $ET1 - ET9$ we see that in addition to zipper sets we only introduce degenerated i·zipper, which are intervals, ($ET4, ET5, ET6, ET8$) and i·zipper sets with non degenerated left side $[a \sim b - c]$, $b = a+2$, $ET7$, and i·zipper sets with non degenerated right side $[b' \sim c' - d]$, $d = c' + 2$, $ET9$. Moreover, the lower and upper bounds $ET7$ and $ET9$ coincide (this comes from $\{j, g\} \prec \{k, h\}$). The statement of the lemma follows from $ET1 - ET9$. Note that $ET3$ must precede all $ETi$ that contain $p$ by Lemma 6. We denote $ET \cup (ET \uplus 1) = \{ETi, ETi \uplus 1) | i = 1, \ldots, 9\}$.

Second, we prove the lemma for $X_{t-1}$. We use a complete enumeration of cases again. This enumeration is feasible due to the restricted framed-structure of $ET \cup (ET \uplus 1)$. Figure 7(c) shows the result of the enumeration (note that we do not show restrictions on the partial order between structures.) In fact, it is sufficient to consider all sets that can be formed as the union of two sets in $ET \cup (ET \uplus 1)$. Let $S = \cup_{s_1, s_2 \in ET \cup (ET \uplus 1)} \{s_1 \cup s_2\}$ be the set of all sets that can be obtained by the union of two sets in $S$. Taking the union of any three sets $s_1, s_2, s_3 \in ET \cup (ET \uplus 1)$ is in $S$ again. So we reach a fixpoint after considering all sets of size two. The resulting structures show that the lemma holds. We do not specify any restriction on the relative order of the framed-structures. □

### C.3 i·Zipper block.

**Lemma 17.** *Consider a* SEQBIN$(N, X, C, B)$ *constraint with monotone $B$ and arbitrary $C$. If $c(x_{n_1+1}, j)$ is an i·zipper set $[a \sim b - c]$ then $b - a \leq 4$. Similarly, If $c(x_{n_1+1}, j)$ is an i·zipper set $[b - c \sim d]$ then $d - c \leq 4$.*

*Proof.* Let $p = \min lb_f(x_{n_1+1}, j)$. There are two ways to obtain i·zipper set $[a \sim b-c]$. Firstly, we can extent an existing i·zipper set. By Lemma 16 we know that for any i·zipper set $[a \sim b - c]$ we have $b = a + 2$. As $a \in \{p, p + 1\}$ for any such i·zipper set we can extend a zipper with at most one hole on the left. For example, we can combine the i·zipper set $NT4$ shifted by one and the zipper $NT11$, $(NT4 \uplus 1) \cup NT11$ to obtain the i·zipper set $[p \sim (p + 4) - (q + 2)]$. Example 4 demonstrates this situation: $c_f(x_{1,3}) = [1 \sim 5 - 8]$. Hence, any extension of an i·zipper set satisfies the lemma.

Secondly, we can create a new i·zipper set. However, as all lower bounds are on distance at most 2 from each other we cannot obtain a new i·zipper set $[a \sim b - c]$ with $b - a > 2$. Hence, the lemma holds. □

**Lemma 18.** *Consider a* SEQBIN$(N, X, C, B)$ *constraint with monotone $B$ and arbitrary $C$. The following hold for all $i \in [1, n_1 + 1]$*

- **Interval monotonicity.** *If $\pi^{top}$ is the top value of $D(X_i)$, then the set $c_f(x_{i, \pi^{top}})$ is an i·zipper set $[a \sim b - c \sim d]$ with $c > b$.*
- **Overlapping intervals.** *If there exist two values $j, k \in D(X_i)$ such that $c_f(x_{i,j})$ and $c_f(x_{i,k})$ are two i·zipper sets $[a \sim b - c \sim d]$ and $[a' \sim b' - c' \sim d']$, respectively, then $[b, c] \cap [b', c'] \neq \emptyset$.*
- **Unique interval.** *$c_f(x_{i,j})$ is an i·zipper set.*
- **Cost structure.** *For any i·zipper set $c_f(x_{i,j}) = [a \sim b - c \sim d]$ it holds $b - a \leq 4$ and $d - c \leq 4$.*

*Proof. Interval monotonicity.* By Lemma 4 we know that the top value $\pi^{top}$, $\pi^{top} \in D(X_{n_1+1})$ contains an interval. Hence, $c_f(x_{n_1+1, \pi^{top}})$ is an i·zipper set. By the monotonicity of $B$, the top value of all $X_i$, $i \in [1, n_1]$ must also have an interval.

*Overlapping intervals.* We show by induction on distance from $n_1 + 1$. The base case was proven in Lemma 16. Now suppose this holds for all $X_{t+1}, \ldots, X_{n_1+1}$. We show it holds for $X_t$. Let $\pi^{top}$ be the top value in $D(X_{t+1})$. By the inductive hypothesis it contains the interval $[a, b], b > a$. By monotonicity of $B$, $\pi^{top}$ supports all values of $X_t$. Therefore, all $c_f(x_{t,j})$ will contain $[a, b]$ or $[a + 1, b + 1]$ for all $j \in D(X_t)$, so all pairs of values of $X_t$ contain intervals that overlap.

*Unique intervals.* We again show this by induction on the distance from $n_1 + 1$. The base case was proven in Lemma 16. Now suppose this holds for all $X_{t+1}, \ldots, X_{n_1+1}$. We show it holds for $X_t$.

Suppose there exists a value $j \in D(X_t)$ such that $c_f(x_{t,j})$ contains two distinct intervals. Consider first any two supports $v$ and $w$ of $j$ in $X_{t+1}$ that contain intervals. By the inductive hypothesis, $c_f(x_{t+1,v})$ contains a unique interval $[a, b]$ and $c_f(x_{t+1,w})$ contains a unique interval $[c, d]$ such that $[a, b]$ and $[c, d]$ overlap. Since $v$ and $w$ support $j$, $c_f(x_{t,j})$ contains $[a, b] + c(j, v)$ which is $[a, b]$ or $[a + 1, b + 1]$ and $[c, d] + c(j, w)$ which is $[c, d]$ or $[c + 1, d + 1]$. Shifting both overlapping intervals by the same amount, i.e., $c(j,v) = c(j,w)$, yields overlapping intervals again, so their union is an interval. If we shift only one, i.e., $c(j,v) \neq c(j,w)$, the intervals will in the worst case be adjacent, therefore their union is an interval. Because the union of any two supporting intervals yields an interval, the union of all supporting intervals also yields an interval. Therefore, if $c_f(x_{t,j})$ contains two distinct intervals, one of them must come from the union of the parts of the supports that form zipper sets.

Assume that $c_f(x_{t,j})$ is of the form $[a \sim b - c \sim d - e \sim f], b < c, c + 1 < d < e$ and, w.l.o.g., assume that the interval $[b,c]$ is the union of zipper sets and the interval $[d,e]$ is the union of intervals.

Since the interval $[b,c]$ is the union of two or more zipper subsets of i·zipper sets, one of the zipper subsets, say $c_f(x_{t+1,v})$, must terminate at $c - 1$, i.e., $ub_f(x_{t+1,v}) = c - 1$, otherwise $c_f(x_{t,j})$ would also contain $c + 1$ and $[b,c]$ would not be a maximal interval. is exactly a zipper set, otherwise it would contain

Moreover, one of the supports of $j$ that support the interval $[d,e]$, say $w$, must be such that $c_f(x_{t+1,w})$ contains the value $e - 1$, so $ub_f(x_{t+1,w}) \geq e - 1 \geq d > c + 1$, so $ub_f(x_{t+1,w}) > c + 1 = (c-1) + 2 = ub_f(x_{t+1,v}) + 2$, or $ub_f(x_{t+1,w}) \geq ub_f(x_{t+1,v}) + 3$ which contradicts Lemma 5.

*Cost structure.* By Lemma 4 and the unique intervals property above we know that any $c(x_{n_1}, j)$ is an i·zipper set and it must contain an interval. By Lemma 17 we know that for any i·zipper $c(x_{n_1}, j) = [a \sim b - c \sim d], j \in D(X_{n_1}), b - a \leq 4$ and $d - c \leq 4$. Hence, lemma holds for the $n_1$th layer. The important observation is that starting from layers $n_1 - 1$ none of $c(x_i, j) = [a \sim b - c \sim d], j \in D(X_t), t \in [1, \ldots, n_1 - 1]$, has a support value at the $(t+1)$th layer whose cost set is a zipper set by Lemma 4.

Suppose the statement holds for $t-1$ layers starting with the $n_1^{th}$ layer. To obtain an i·zipper set $c_f(x_t, j)$ we must combine two support values $v, w \in D(X_{t+1})$ such that $c_f(x_{t+1}, u) = [a_u \sim b_u - c_u \sim d_u], u \in \{v, w\}, b_u - a_u \leq 4$ and $d_u - c_u \leq 4$ as there are only available framed-structures of $c(x_{t+1}, u), u \in D(X_{t+1})$ that contain holes. As $[b_v, c_v] \cap [b_w, c_w] \neq \emptyset, |a_v - a_w| \leq 2$ and $|d_v - d_w| \leq 2, b_v - a_v \leq 4, d_v - c_v \leq 4$ $b_w - a_w \leq 4$ and $d_w - c_w \leq 4$, the union of any subset of $c_f(x_{t+1}, j) \cup (c_f(x_{t+1}, j) + 1)$ can not create an i·zipper set $[a' \sim b' - c' \sim d']$ such that $b - a > 4$ or $d - c > 4$. Hence, the lemma holds for the $t^{th}$ layer. □

**Corollary 6.** *The number of distinct sets of $c_f(x_{i,j})$, $j \in D(X_i)$, is bounded at the ith layer, $i = 1, \ldots, n$.*

*Proof.* Consider the set $c_f(x_{i,j}), j \in D(X_i)$. By Lemma 15 each variable in the zipper block of variables contains at most 4 different framed-structures of costs. By Lemma 16 each variable in the zipper + i·zipper block of variables contains at most 18 different framed-structures of costs. By Lemma 18 each variable in the i·zipper block of variables contains at most 9 different i·zipper sets. Each of them can start/finish in at most 3 points. This gives at most 81 possible distinct framed-structures. Hence, the number of distinct sets of $c_f(x_{i,j}), j \in D(X_i)$ is bounded. □

## D Domain consistency algorithm (omitted proofs)

**Lemma 19.** *Consider a* SEQBIN$(N, X, C, B)$ *constraint such that $B$ is monotone. For for all $j \in D(X_i), i \in [1, \ldots, n]$ $c_f(x_{i,j}) = \bigcup_{v \in D(X_{i+1})} (c_f(x_{i+1,v}) \uplus c(j, v))$ can be computed in $O(d)$ time.*

*Proof.* We partition all supports $v$ into two groups based on the value of $c(j, v)$. The first group $S_0$ contains values such that $c(j, v) = 0$ and the second group $S_1$ contains values such that $c(j, v) = 1$. We find $c^1 = \bigcup_{v \in S_0} c_f(x_{i+1,v})$ and $c^2 = \bigcup_{v \in S_1} c_f(x_{i+1,v})$. Then we find $c_f(x_{i,j}) = c^1 \cup (c^2 \uplus 1)$. We prove lemma for $c^1$ ($c^2$ is analogous.)

*Compute $c^1$.* We assume that $p$ is the smallest lower bound among the forward cost sets of the values in $S_0$ and $q + 2$ is the greatest upper bound: $p = \min_{v \in S_0} lb_f(x_{i+1,v})$ and $q + 2 = \max_{v \in S_0} ub_f(x_{i+1,v})$. We refer to $l \cdot zip$ of $c_f(x_{i+1,v})$ as $l \cdot zip(x_{i+1,v})$ to simplify notations (similarly, for the other two parts $i \cdot val$ and $r \cdot zip$). By Theorem 1 we know that $lb(l \cdot zip(x_{i+1,v})) \in [p, p+2]$, $ub(r \cdot zip(x_{i+1,v})) \in [q, q+2]$, $lb(i \cdot val(x_{i+1,v})) \in [p, p+6]$ and $ub(i \cdot val(x_{i+1,v})) \in [q-4, q+2]$. Hence, we compute the 20 indicator values $J_y^{l \cdot zip}(v), J_y^{r \cdot zip}(v), J_y^{i \cdot val_{lb}}(v)$ and $J_y^{i \cdot val_{ub}}(v), v \in S_0$. We define $J_y^{l \cdot zip}(v) = 1$, iff $lb(l \cdot zip(x_{i+1,v})) = y$ and $J_y^{l \cdot zip}(S_0) = \max_{v \in S_0} J_y^{l \cdot zip}(v)$. Similarly, we compute the other 19 indicators. This can be done in $O(d)$ time with a linear scan over $c_f(x_{i+1,v}), v \in S_0$.

Then we can compute $\bigcup_{v \in S_0} c_f(x_{i+1,v}) = [a^* \sim b^* - c^* \sim d^*]$ in 4 steps, each of which takes $O(1)$ time.

*Union of $i \cdot val$.* Theorem 1 shows that all $i \cdot val$ sets must overlap. Hence, the union of $i \cdot val(x_{i+1,j})$ forms an interval. We find the minimum value $y$, $y \in \{p, \ldots, p+6\}$ such that $J_y^{i \cdot val_{lb}}(S_0) = 1$. If such a value $y$ exists then we set $b^* = y$. Then we find the largest value $y' \in \{q-4, \ldots, q+2\}$ such that $J_y^{i \cdot val_{ub}}(S_0) = 1$ and set $c^* = y'$. Note that if $y$ exists then $y'$ exists. If $y$ does not exist we know that all $c_f(x_{i+1,v}), v \in S_0$ are zipper sets and we set $b^* = c^* = \emptyset$.

*Union of $l \cdot zip$.* Suppose $b^* \neq \emptyset$. We find indicators $J_y^{l \cdot zip}(S_0), y \in [p, p+2]$, that are set to one. Set $p'$ to the minimum among $[p, p+2]$, for which there exists $J_{p'}^{l \cdot zip}(S_0) = 1$. If $J_{p+1}^{l \cdot zip}(S_0) = 1$ and $J_p^{l \cdot zip}(S_0) = 1$ or $J_{p+2}^{l \cdot zip}(S_0) = 1$ or $b^* \in \{p, p+2\}$ then set $a^*$ and reset $b^*$ $a^* = b^* = \min(p+1, p')$ otherwise set $a^* = p'$ and leave $b^*$ unchanged. Union of $r \cdot zip$ is similar to union of $l \cdot zip$.

*Union of zippers.* Suppose $b^* = \emptyset$. Then we determine which of 4 distinct sets (Lemma 15) are presented among $c_f^1(x_{i+1,v}), v \in S_0$. As there are at most 4 such that are zippers we can union them in $O(1)$ time and identify values of $a^*, b^*, c^*$ and $d^*$.

*Compute $c^1 \cup (c^2 \uplus 1)$ .*

The $\uplus$ operation takes $O(1)$ time. We denote $c'^2 = (c^2 \uplus 1)$. We show that we perform $c^1 \cup c'^2$ in $O(1)$. We only consider the case when $c^1$ and $c'^2$ are i·zipper, $c^1 = [a \sim b - c \sim d], c'^2 = [(a'+1) \sim (b'+1) - (c'+1) \sim (d'+1)]$. Reasoning about other cases is analogous.

We construct the union in three steps $c_f(x_{i,j}) = [a^* \sim b^* - c^* \sim d^*] = c''^1 \cup c''^2 \cup c''^3$, where $c''^1 = [a \sim b] \cup [(a'+1) \sim (b'+1)]$, $c''^2 = [b, c] \cup [(b'+1), (c'+1)]$ and $c''^3 = [c \sim d] \cup [(c'+1) \sim (d'+1)]$.

By Theorem 1 we know that $[b, c] \cap [b', c'] \neq \emptyset$. Hence, $[b, c]$ and $[b' + 1, c' + 1]$ intersect or touch so that $c = b' - 1$ if $b \leq b'$. In any case $c''^2 = [b, c] \cup [(b' + 1), (c' + 1)] = [b^*, c^*]$ is an interval, where $b^* = \min(b, (b'+1))$ and $c^* = \max(c, (c'+1))$. Next we consider the $l \cdot zip$ parts of $c^1$ and $c'^2$, $[a \sim b]$ and $[(a'+1) \sim (b'+1)]$, respectively. It suffices to restrict these sets to $[a \sim b^*]$ and $[a' \sim b^*]$ as $b^* = \min(b, (b'+1))$. Note that $c''^1 = [a \sim b^*] \cup [(a'+1) \sim b^*]$ can be either an interval or a zipper and an i·zipper of the form $[r \sim q - b^*]$ depending on the parities of $a$ and $a'$. In any case, by Theorem 1, $b^* - a \leq 4$, $b^* - (a'+1) \leq 4$, hence there is a bounded number of outcomes and we compute the union in constant time. Similarly, we compute $c''^3$ which is the union of $[c^* \sim d]$ and $[c^* \sim (d'+1)]$. Hence, we can compute the union $c^1 \cup c'^2$ in $O(1)$ time.

*Complexity analysis* . For each forward cost set $c_f(x_{i,j})$, $j \in D(X_i)$, $i \in [1, \ldots, n]$ can be computed in $O(d)$. As we have $O(nd)$ forward cost sets, the total time complexity is $O(nd^2)$. Note that one way to reduce this complexity is to compute $c_f(x_{i,j})$ in $O(1)$.

□